\documentclass[conference,compsoc]{IEEEtran}
\ifCLASSOPTIONcompsoc
  \usepackage[nocompress]{cite}
\else
  \usepackage{cite}
\fi
\ifCLASSINFOpdf
\else
\fi
\usepackage{amsmath}
\usepackage{url}

\usepackage[table]{xcolor}
\usepackage{graphicx}
\usepackage{xspace}
\usepackage{subfig}
\usepackage{booktabs}
\usepackage{multirow}

\newcommand{\system}{DiffSign\xspace}

\begin{document}
%
\title{T2I-Based Physical-World Appearance Attack against Traffic Sign Recognition Systems in Autonomous Driving}

\author{{\rm Chen Ma}$^{\star}$  \quad {\rm Ningfei Wang}$^{\dagger}$ \quad {\rm Junhao Zheng}$^{\star}$ \quad {\rm Qing Guo}$^{\ddagger}$ \quad {\rm Qian Wang}$^{\mathparagraph}$ \quad {\rm Qi Alfred Chen}$^\dagger$ \quad {\rm Chao Shen}$^{\star}$ \\
$^\star$Xi'an Jiaotong University \quad
$^\dagger$University of California, Irvine \quad
$^\ddagger$Nankai University \quad
$^{\mathparagraph}$Wuhan University
}

\maketitle

\begin{abstract}
Traffic Sign Recognition (TSR) systems play a critical role in Autonomous Driving (AD) systems, enabling real-time detection of road signs, such as STOP and speed limit signs. While these systems are increasingly integrated into commercial vehicles, recent research has exposed their vulnerability to physical-world adversarial appearance attacks. In such attacks, carefully crafted visual patterns are misinterpreted by TSR models as legitimate traffic signs, while remaining inconspicuous or benign to human observers. However, existing adversarial appearance attacks suffer from notable limitations. Pixel-level perturbation-based methods often lack stealthiness and tend to overfit to specific surrogate models, resulting in poor transferability to real-world TSR systems. On the other hand, text-to-image (T2I) diffusion model-based approaches demonstrate limited effectiveness and poor generalization to out-of-distribution sign types.

In this paper, we present \system, a novel T2I-based appearance attack framework designed to generate physically robust, highly effective, transferable, practical, and stealthy appearance attacks against TSR systems. To overcome the limitations of prior approaches, we propose a carefully designed attack pipeline that integrates CLIP-based loss and masked prompts to improve attack focus and controllability. We also propose two novel style customization methods to guide visual appearance and improve out-of-domain traffic sign attack generalization and attack stealthiness. We conduct extensive evaluations of \system under varied real-world conditions, including different distances, angles, light conditions, and sign categories. Our method achieves an average physical-world attack success rate of 83.3\%, leveraging \system's high effectiveness in attack transferability.

\end{abstract}


%

\section{Introduction}
\label{sec:intro}
Traffic Sign Recognition (TSR) systems are a critical component of Advanced Driver Assistance Systems (ADAS), responsible for detecting and interpreting road signs such as speed limit signs and STOP signs using onboard camera sensors~\cite{akatsuka1987road, kbb_tsr}. In recent years, TSR systems typically rely on Deep Neural Networks (DNNs) for real-time traffic sign detection due to their superior performance~\cite{ravindran2019traffic, alawaji2024traffic, wang2025revisiting}. As a result, TSR systems powered by DNNs have been widely deployed across a broad range of commercial vehicles~\cite{wang2025revisiting, sato2024intriguing, sato2024invisible}. Given the safety-critical nature of traffic sign interpretation—where failures to correctly recognize or respond to signage can lead to severe consequences such as traffic rule violations or accidents—the security of TSR systems has become a pressing concern. A growing body of research has thus focused on the security of these systems, particularly under real-world conditions~\cite{zhao2019seeing, sato2024intriguing, jia2022fooling, nassi2020phantom, eykholt2018physical, wang2025revisiting, zhu2023tpatch, Wang_2023_ICCV, he2024dorpatch}.

\begin{figure}[!t]
\centering
\subfloat[SIB~\cite{zhao2019seeing}]
{\includegraphics[height=0.21\linewidth]
{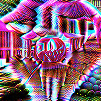}
\label{fig_1a}
}
\subfloat[FTE~\cite{jia2022fooling}]
{\includegraphics[height=0.21\linewidth]
{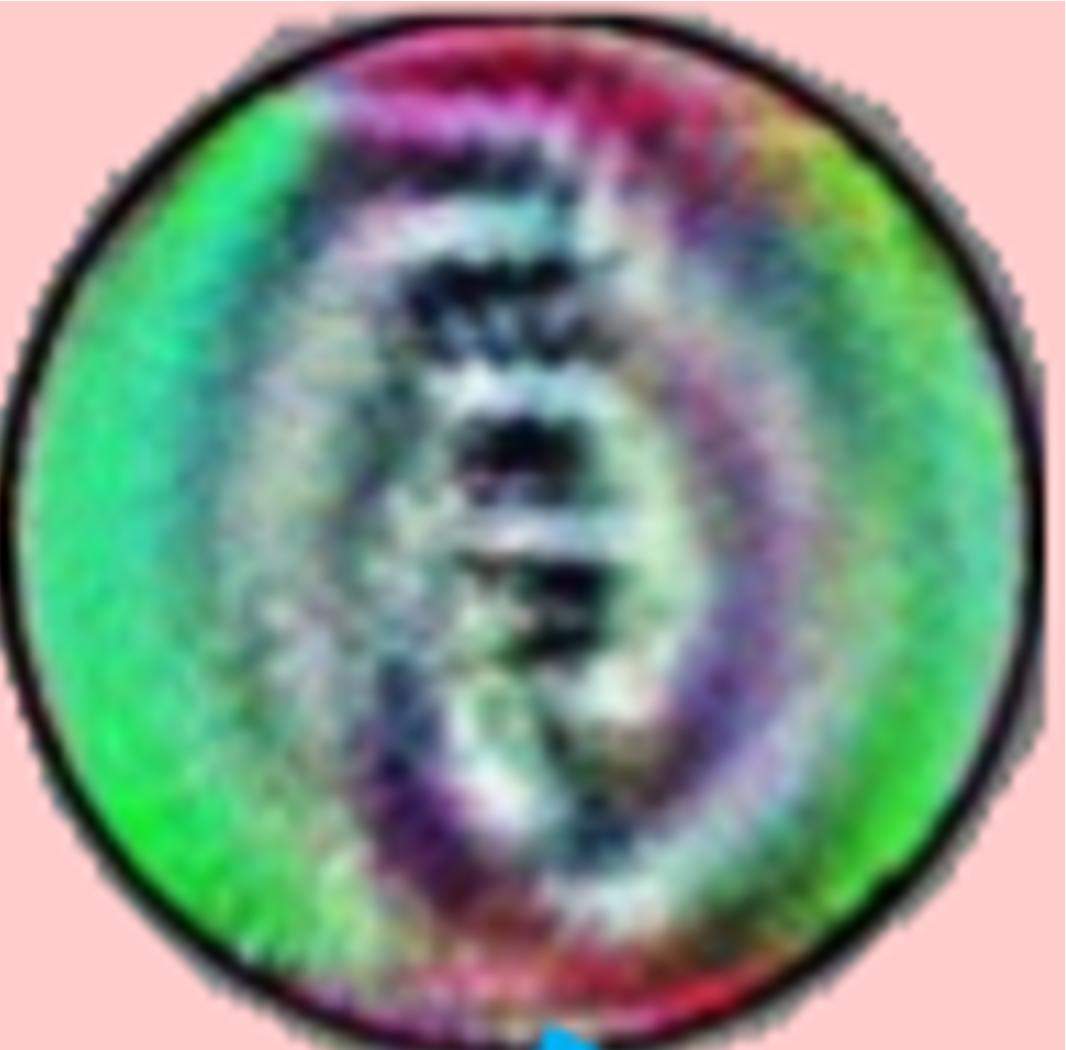}
\label{fig_1c}
}
\subfloat[\system(ours)]
{\includegraphics[height=0.21\linewidth]
{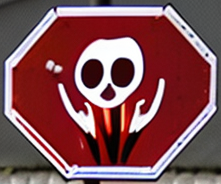}
\label{fig_1b}
}
\subfloat[\system(ours)]
{\includegraphics[height=0.235\linewidth]
{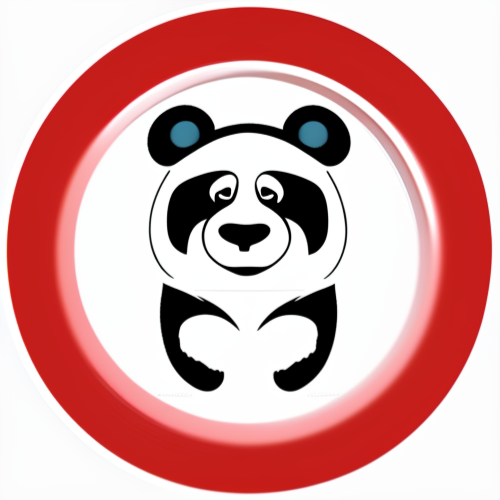}
\label{fig_1d}
}
\caption{Visualization of representative existing TSR attacks: SIB~\cite{zhao2019seeing} on STOP sign and FTE~\cite{jia2022fooling} on speed limit sign, compared with our newly proposed attack \system, respectively.}
\vspace{-0.5cm}

\label{fig:example}

\end{figure}

Among the various attacks on TSR systems~\cite{zhao2019seeing, sato2024intriguing, jia2022fooling, nassi2020phantom, eykholt2018physical}, physical-world appearance attacks are particularly alarming,
which makes the TSR systems misdetect an object as a specific target traffic sign chosen by the attacker, while remaining imperceptible to human observers. Unlike other attack types that require direct access to existing traffic signs to attach adversarial patches or stickers, making them location-dependent and easily detectable through routine inspections, appearance attacks present a far more severe and flexible threat. Appearance attacks can be deployed through diverse and inconspicuous carriers that naturally appear in driving scenarios. Specifically, adversarial examples can be placed on roadside billboards, printed on the rear panels of vehicles traveling ahead of the victim, or even disguised as clothing patterns worn by pedestrians near the roadside. This flexibility makes appearance attacks not only harder to detect and prevent but also significantly more scalable, as attackers can launch attacks without physically tampering with legitimate traffic infrastructure.
Such attacks pose severe safety risks in Autonomous Driving (AD) scenarios. For instance, an adversarial example perceived as a 5 km/h speed limit sign on a highway may trigger sudden deceleration of the AD vehicle, potentially causing rear-end collisions. Similarly, misdetection of a STOP sign can result in unnecessary emergency stops in fast-moving traffic.

However, existing appearance attacks exhibit several notable limitations. 
Traditional pixel-level attacks, such as FTE~\cite{jia2022fooling} and SIB~\cite{zhao2019seeing}, suffer from poor naturalness and stealthiness in the generated adversarial examples, as illustrated in Fig.\ref{fig:example} (a) and (b). The resulting examples contain visually noticeable noise artifacts that can easily arouse suspicion from human observers, undermining the covert nature required for practical deployment. For instance, deploying such conspicuous examples on billboards would likely attract scrutiny from law enforcement and regulatory authorities. These methods generate adversarial appearance attacks by applying fine-grained pixel-level perturbations, which inherently limits their control over the visual characteristics of the generated patterns. Furthermore, pixel-level methods tend to overfit to the surrogate or source model used during attack generation, which severely limits their transferability to unseen models, as evaluated in~\S\ref{sec:attack_effectiveness}. This low transferability significantly reduces their ability to attack commercial TSR systems effectively~\cite{wang2025revisiting}, thereby limiting their practical real-world impact.

To further improve the generated attack patterns, the state-of-the-art appearance attack method NDD~\cite{sato2024intriguing} employs text-to-image (T2I) models to generate natural-looking adversarial examples, which achieves high attack stealthiness but suffers from generally low attack effectiveness (achieving only 15.3\% attack success rate), as evaluated in~\S\ref{sec:attack_effectiveness}. This limitation stems from three fundamental challenges. \textbf{First}, poor generalizability, as attack capability of NDD is restricted to STOP signs and fails to extend to other sign types. This limitation arises because T2I-based methods such as NDD heavily rely on the prior knowledge embedded in the diffusion model, which is inherently constrained by the categories present in its training dataset. Consequently, these models struggle with out-of-domain objects that lack sufficient prior knowledge, such as Chinese speed limit signs or region-specific regulatory signs. \textbf{Second}, NDD lacks explicit adversarial guidance during the generation process, preventing it from effectively focusing on the primary attack target within the full image generated by the T2I model. This absence of targeted optimization results in diffuse adversarial patterns that fail to concentrate their attack potential on the intended object, leading to significantly reduced attack effectiveness. \textbf{Third}, T2I-based appearance attacks suffer from poor controllability, often producing adversarial examples that unintentionally retain recognizable semantic features from benign signs. For instance, generated patterns frequently contain readable text elements like "STOP" or numeric indicators such as "5", making them easily identifiable to human observers and thus disqualifying them as legitimate appearance attacks. Moreover, these semantic artifacts make the attacks vulnerable to detection by semantic-aware defense methods such as ContraNet~\cite{Yang2022WhatYS}.

To address these limitations, we propose a novel attack method called \system, which generates physical-world adversarial appearance attacks with semantics using T2I diffusion models. Specifically, the newly proposed generation framework involves a cropping mechanism and bounding box filtering to enhance the attack focusing on the primary object. Moreover, we design masked prompts and a new loss function based on the CLIP model~\cite{radford2021learning}, which further improves the controllability of the T2I process and enhances the qualification of generated examples. Furthermore, we introduce two style customization techniques: image-specified customization, which mimics the appearance of attacker-provided reference images; and prompt-specified customization, which modifies signs using natural language guidance. The style customization can guide the T2I process to focus on controlled semantics. This significantly enhances \system’s capability for generating out-of-domain adversarial attacks. Additionally, controlling the attack style, such as its shape and texture, can further improve the stealthiness of the attacks~\cite{zhao2019seeing, song2018physical}. Existing style customization methods often fail to achieve precise control, resulting in outputs that appear noisy~\cite{zhao2019seeing, song2018physical} or deviate from the intended styles~\cite{sato2024intriguing}.

Compared to traditional pixel-based attacks~\cite{zhao2019seeing, jia2022fooling}, \system offer several main advantages over pixel-level physical-world attacks. \textit{Enhanced Stealthiness:} By leveraging the semantic information captured by T2I diffusion models, \system produces adversarial attacks that are more visually natural and less noticeable to humans. The generated attacks maintain high perceptual quality and are considered likely to appear in our daily lives, as supported by our user study results detailed in~\S\ref{sec:attack_effectiveness}. \textit{Improved Robustness and Transferability}: These examples demonstrate excellent robustness in physical-world conditions and stronger transferability. Our appearance attack successfully attacks the two commercial TSR systems of two leading production vehicle brands with two sign types, achieving a 97\% attack success rate. \textit{Higher Attack Efficiency}: We first discover that T2I-based methods for physical adversarial attacks require fewer Expectation-over-Transformation (EoT) iterations to improve attack robustness, resulting in lower time costs and higher efficiency.

We evaluate \system under various physical-world conditions, including different angles, distances, light conditions, and sign types, achieving an overall average attack success rate of 83.3\%. To the best of our knowledge, this work represents the first comprehensive and systematic evaluation of T2I-based adversarial attacks under diverse physical-world conditions such as viewing angles, distances, light conditions, and sign types. We also compare it with representative baseline attacks, including NDD~\cite{sato2024intriguing} and SIB~\cite{zhao2019seeing}. The results demonstrate that our attack significantly outperforms existing methods, achieving approximately 3.6 times higher success rate than NDD  and 4.9 times higher than SIB.

\system shows strong attack transferability across seven TSR models. Furthermore, \system demonstrates high attack transferability effectiveness on two commercial TSR systems in leading production vehicle brands, achieving an overall 97\% success rate across different sign positions and sign types. Additionally, to understand the end-to-end impact of the attack, such as unnecessary emergency stops, we evaluate the system-level impact using the \textit{PASS}~\cite{shen2022sok, hu2022pass}. Our \system triggers unnecessary emergency stops with 100\%, compared to a 0\% success rate for baseline attacks. We evaluate both general DNN-based defenses and domain-specific defense strategies, and propose promising directions for future defense development. We discuss the limitations of our work, outline potential avenues for future research, and address the ethical considerations associated with this study.

To sum up, this paper makes the following contributions:
\begin{itemize}
\item We propose a novel T2I-based physical-world adversarial appearance attack \system against TSR systems by overcoming limitations of existing pixel-level and T2I-based attacks.
\item  To enhance attack focus and controllability, we introduce a cropping mechanism, CLIP-based loss, and masked prompts. We also propose two novel style customizations to guide visual appearance and improve out-of-domain traffic sign attack generalization and attack stealthiness.
\item  We conduct an extensive evaluation of our attack across various aspects, including effectiveness, stealthiness, attack transferability, and system-level impact on AD systems. The results demonstrate that \system is highly effective, stealthy, transferable, and provides superior style customization compared to previous attacks.
\item \system can generate transferable physical-world attacks, exhibiting excellent effectiveness against two commercial TSR systems in the leading production vehicle brands with different sign types and sign locations.
\end{itemize}

\section{Background and Related Work}

\label{sec:background}
\subsection{Traffic Sign Recognition (TSR) System}
\label{sec:TSR_system}
Traffic sign recognition (TSR) systems utilize camera sensor outputs as inputs for DNN to detect the traffic signs such as STOP sign and speed limit sign by performing real-time object detection~\cite{almutairy2019arts, sato2024invisible, wang2025revisiting}. These systems are not only capable but also cost-effective, contributing to their widespread integration into AD systems, including those developed by companies~\cite{rezgui2019traffic}. State-of-the-art object detectors employed in these systems fall into two primary categories: one-stage and two-stage object detectors~\cite{Wang_2023_ICCV}. One-stage detectors, exemplified by YOLO~\cite{redmon2018yolov3}, are known for their high detection speed. On the other hand, two-stage detectors, such as Faster R-CNN~\cite{ren2015faster}, are recognized for their superior detection accuracy. Our work includes object detectors from both types to provide a comprehensive analysis of TSR systems.

\subsection{Text-to-Image (T2I) Diffusion Model}
\label{sec:diffussion_model}
Denoising Diffusion Probabilistic Models (DDPMs)~\cite{ho2020denoising} have demonstrated remarkable success in generating high-quality images through multi-step Markov chain simulations~\cite{nichol2021improved, dhariwal2021diffusion}. These models operate in two main stages: forward and reverse diffusion. In the forward process, a clean image is progressively corrupted with Gaussian noise, while in the reverse process, the model is trained to iteratively remove this noise, effectively reconstructing the original image. Despite its simplicity, this approach achieves impressive image quality. To enhance efficiency, the Denoising Diffusion Implicit Model (DDIM)\cite{song2020denoising} is introduced, using a similar training framework as DDPMs but optimized for faster image generation. A further extension, text-to-image diffusion models, enables image generation guided by text prompts. This capability has led to widespread use in major models such as Stable Diffusion~\cite{rombach2022high}, where contrastive image-text models such as CLIP~\cite{radford2021learning} are incorporated to provide text guidance during both training and inference. Given the diffusion model’s ability to improve adversarial effectiveness, transferability, robustness, and stealthiness~\cite{chen2024diffusion, sato2024intriguing}, we propose a novel attack framework, \system, to generate physical-world adversarial appearance attacks for TSR systems.

\subsection{Appearance Attack on TSR Systems}
\label{sec:adversarial_attack}

Physical-world appearance attacks have emerged as a particularly compelling 
class of threats to TSR systems~\cite{eykholt2018physical, zhao2019seeing, 
jia2022fooling, sato2024intriguing}. These attacks exploit the discrepancy 
between features learned by deep models and those used by human visual 
perception. By generating adversarial patches and placing them near road, attackers can cause TSR systems to misrecognize 
them as specific target signs, while ensuring these patches are not recognized as traffic signs by human observers.
Early pixel-based methods~\cite{eykholt2018physical, 
zhao2019seeing, jia2022fooling} optimize individual pixel values to generate 
adversarial patches, resulting in conspicuous, unnatural patterns that may 
alert human drivers or raise legal concerns. Recent work has explored 
diffusion-based text-to-image models to generate more natural-looking 
patches~\cite{sato2024intriguing}, but these approaches demonstrate limited 
effectiveness under physical-world conditions.
In this paper, we present \system, a novel physical-world appearance attack 
that overcomes both challenges. \system generates adversarial patches that 
mimic common roadside elements such as advertisement logos or graffiti—patterns 
that are inconspicuous to drivers and commonly seen in urban environments, 
thus unlikely to draw attention. At the same time, \system achieves 
significantly higher attack success rates than prior work, demonstrating 
strong effectiveness against commercial AD systems in real-world scenarios 
(\S\ref{sec:commercial}), while maintaining high stealthiness as validated 
by our user study (\S\ref{sec:user-study}).

\section{Threat Model and Design Challenges}

\begin{figure}[!t]
\centering

{\includegraphics[width=1.0\linewidth]
{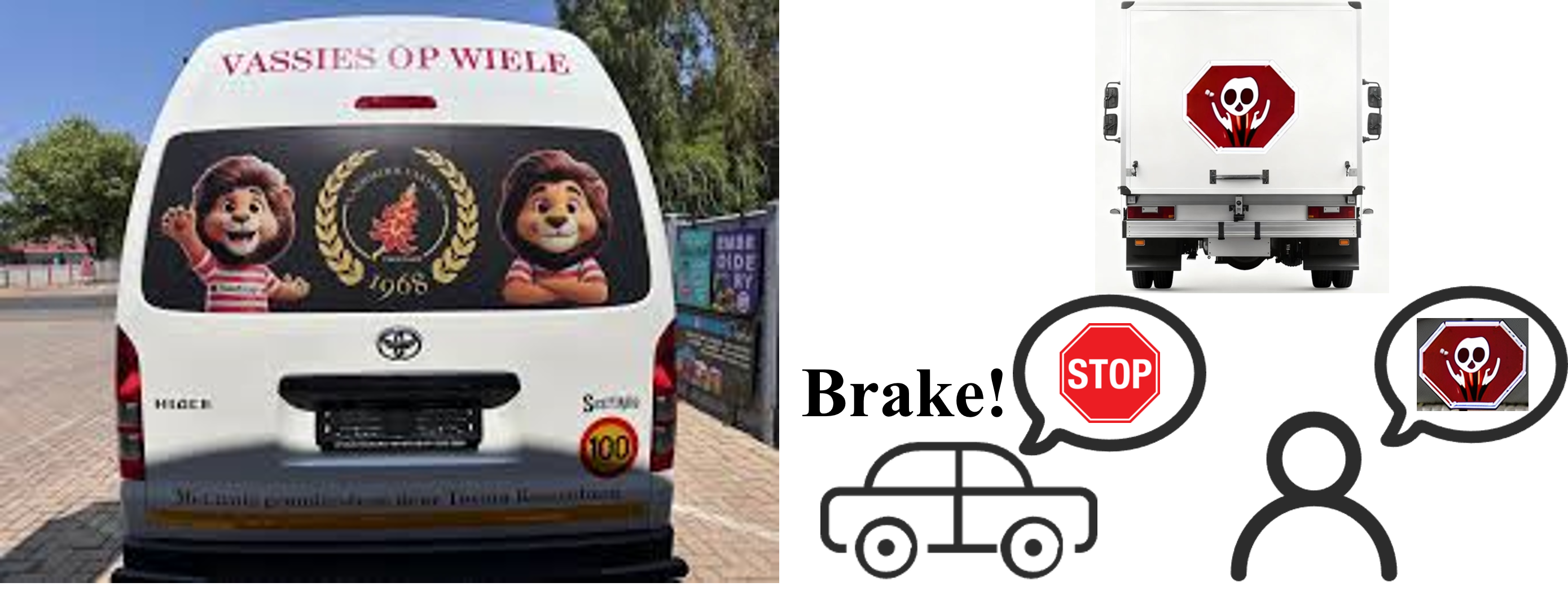}
}

\caption{Example of our attack scenario in the real world (Left) and a schematic of the appearance attack (Right).}

\vspace{-0.3cm}
\label{fig:threat-model}

\end{figure}

\subsection{Threat Model}
\label{sec:threat_model}

\begin{figure*}[!t]
\centering
\includegraphics[width=1.0\linewidth]
{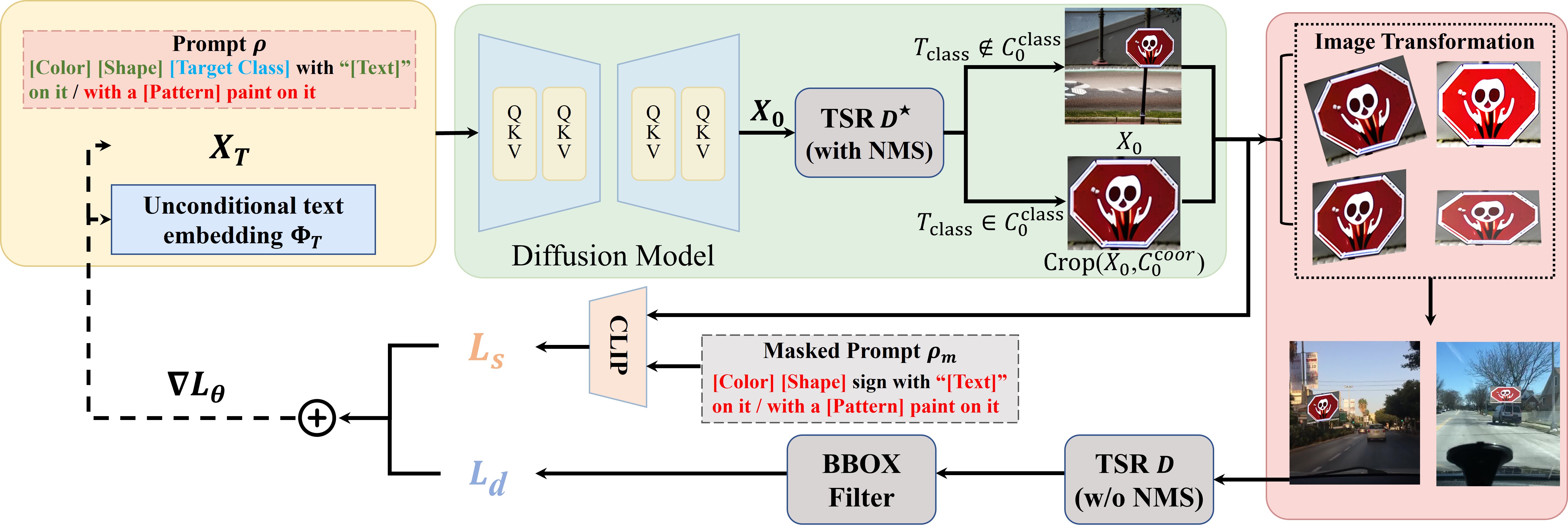}

\vspace{0.1cm}
\caption{Design overview of our attack, \system, a T2I-based physical-world appearance attack.}
\vspace{-0.2cm}
\label{fig:overview}

\end{figure*}

The attack goal is to create an adversarial example that appears as natural elements—such as advertisement logos or graffiti art, while being misrecognized as a specific traffic sign by TSR systems as shown in Fig.~\ref{fig:threat-model}. This offers three advantages: (1) \textit{Legal plausibility}: unlike directly 
defacing existing traffic signs or fabricating 
sign-like images, our patches resemble 
ordinary urban visual elements that do not violate traffic regulations; 
(2) \textit{Reduced suspicion}: natural-looking patterns such as commercial logos or street art are ubiquitous in urban environments and unlikely to raise alerts from human drivers or authorities. (3) \textit{Flexibility in Deployment}: The natural appearance of our adversarial example allows it to seamlessly blend into everyday elements, such as patterns on T-shirts or advertisements and stickers on the back of vehicles as shown in Fig.~\ref{fig:threat-model}.
We begin by designing a white-box attack, assuming the attacker has full knowledge of the TSR system within the victim AD system—a common threat model in prior AD security research~\cite{cao2021invisible, sato2021dirty, Wang_2023_ICCV, ma2024slowtrack, eykholt2018physical, zhao2019seeing, jia2022fooling, ma2025controlloc, ma2023wip}. While our method is developed under this white-box assumption, it is inherently extensible to a black-box threat model by exploiting the transferability of adversarial patterns. Given the greater practicality of black-box settings, we further evaluate our attack in two ways: (1) by assessing transferability across open-source models for TSR (\S\ref{sec:attack_effectiveness}), and (2) through direct black-box testing on the TSR systems in commercial vehicles (\S\ref{sec:commercial}). DiffSign demonstrates high attack transferability effectiveness on two commercial TSR systems in leading production vehicle brands, indicating its strong black-box potential.

\subsection{Design Challenges}
\label{sec:challenge}

Applying text-to-image (T2I) diffusion models in physical-world appearance attacks on TSR~\cite{sato2024intriguing} meets three main challenges not addressed in previous research, which are outlined below.

{\it C1: Focus on the Primary Object.} T2I diffusion models are designed to generate entire scenes, often blending foreground objects with background regions~\cite{chen2024anyscene, song2023objectstitch, farshad2023scenegenie}. Without explicit localization, the optimization process for attack generation may distribute adversarial patterns across irrelevant areas such as the background of the scene, where the attacker does not intend to target or has difficulties in launching the attacks. This will damage attack effectiveness and produce invalid appearance attacks. A representative example can be found in Fig.~\ref{fig:bad_case} (a), where the appearance attack is generally not located at the desired object location. Furthermore, another failure mode occurs when the optimization process focuses only on partial regions of the primary object, as illustrated in Fig.\ref{fig:bad_case} (b), which results in a significantly reduced attack success rate of merely 19.2\% (as detailed in~\S~\ref{sec:ablation}).

{\it C2: Controllability in Attack Generation for Attack Qualification.} Diffusion-based T2I often suffers from poor controllability, which compromises the qualification of appearance attacks. For example, NDD~\cite{sato2024intriguing} relies solely on text prompts to guide generation. However, when generating STOP sign or speed limit sign appearance attacks, these prompts frequently result in adversarial attacks that retain critical features from the original benign signs—such as the word ``STOP'' or numeric indicators like ``5'', making them easily identifiable to human observers and thus disqualified as legitimate appearance attacks. This limitation is empirically supported by our user study~\S~\ref{sec:user-study}, where 68\% of participants correctly identified NDD-generated adversarial examples as benign traffic signs, indicating a notable failure in concealing key semantic features. In summary, weak control over the generated content results in adversarial attacks that unintentionally preserve key visual cues from benign traffic signs, undermining their stealth and qualification.

{\it C3: Generalization to Out-of-Domain Traffic Sign Categories.} T2I-based attacks strongly rely on the T2I diffusion model’s prior knowledge of the target object category. As a result, their effectiveness drops sharply when attacking out-of-domain objects—such as Chinese traffic signs—that are absent from the T2I diffusion model’s training distribution. This limitation is quantitatively demonstrated in our ablation study~\S\ref{sec:ablation}, where attacking the in-domain category STOP sign achieves an attack success rate of 82.2\%, but drops drastically to only 14.6\% when applied to the out-of-domain category Chinese speed limit sign.
Thereby, heavy reliance on prior knowledge limits attack generality, making it relatively ineffective for unfamiliar sign types.

\section{Attack Design: \system}
\label{sec:method}

In this section, we detail our attack, named \system, which is a T2I-based physical-world appearance attack, addressing the three main challenges mentioned in~\S\ref{sec:challenge}.

\subsection{Design Overview} 

{\it Enhancing Attack Focus.} To address C1, we isolate the primary object during optimization using bounding box filtering and the object cropping mechanism. This ensures precise spatial alignment between the adversarial patterns and the primary object, and the optimization process focuses on the primary object instead of any other locations, which can significantly improve both effectiveness and efficiency. 

{\it Strengthening Generation Controllability for Attack Qualification.} To address C2, i.e, ensuring generated attacks reflect attacker intent and avoid retaining features from benign signs, which denotes attack qualification, we adopt masked prompts and CLIP-based guidance with loss function designs. This aligns the generation process with desired semantic features while suppressing unintended benign elements.

{\it Improving Generalization to Out-of-Domain Categories.} To address C3, we introduce two style customization techniques: image-specified customization, which mimics the appearance of attacker-provided reference images; and prompt-specified customization, which modifies signs using natural language guidance. The style customization can guide the diffusion process to focus on controlled semantics. This significantly enhances \system’s capability for generating out-of-domain adversarial attacks. Additionally, controlling the attack style, such as its shape and texture, can further improve the stealthiness of the attacks~\cite{zhao2019seeing, song2018physical}. Existing style customization methods often fail to achieve precise control, resulting in noisy outputs~\cite{zhao2019seeing, song2018physical} or deviating from the intended styles~\cite{sato2024intriguing}.

\subsection{\system Methodology Overview}
The overview of our \system is presented in Fig.~\ref{fig:overview}.
The input components for the diffusion model in Fig.~\ref{fig:overview} include three main parts: the condition embedding $\eta=\psi(\rho)$ of prompts $\rho$ ($\psi$ is text encoder), the latent variable $X_T$, and the unconditional text embedding $\Phi_T$, where $T$ represents the number of steps to generate the new images.

The prompt consists of several key elements such as shape, color, and texture (e.g., text and patterns). In the context of TSR, features such as the shape, color, and texture (e.g., the word ``STOP'' on a STOP sign) are critical for the human visual system (HVS)~\cite{ge2022contributions, grill2004human, sato2024intriguing}, as these enable humans to quickly and accurately recognize traffic signs. To enhance the attack's stealthiness, we opt to remove or modify these features that are most significant to human recognition. An example is shown in Fig.~\ref{fig:example} (c), where we use ``with a ghost paint on it'' to describe the texture in the prompt to generate the attack with a ghost pattern to trick the TSR system into detecting it as a STOP sign. We refer to this modified section of the prompt as the ``adversarial prompt'' (red font in Fig.~\ref{fig:overview}). The green and blue font in Fig.~\ref{fig:overview} represents the features of the benign and the target class.

\begin{figure}[!t]
\centering
\subfloat[w/o crop]
{\includegraphics[height=0.30\linewidth]
{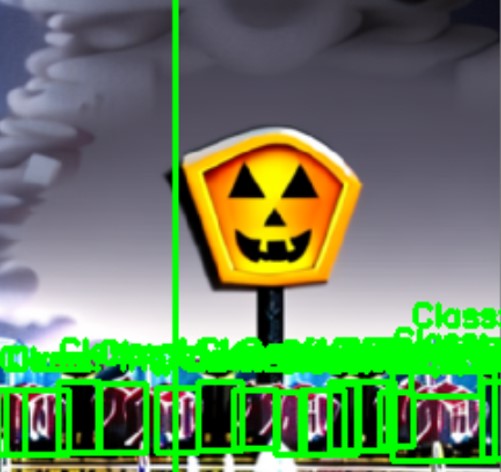}
\label{fig_3a}
}
\subfloat[w/o BBOX filter]
{\includegraphics[height=0.30\linewidth]
{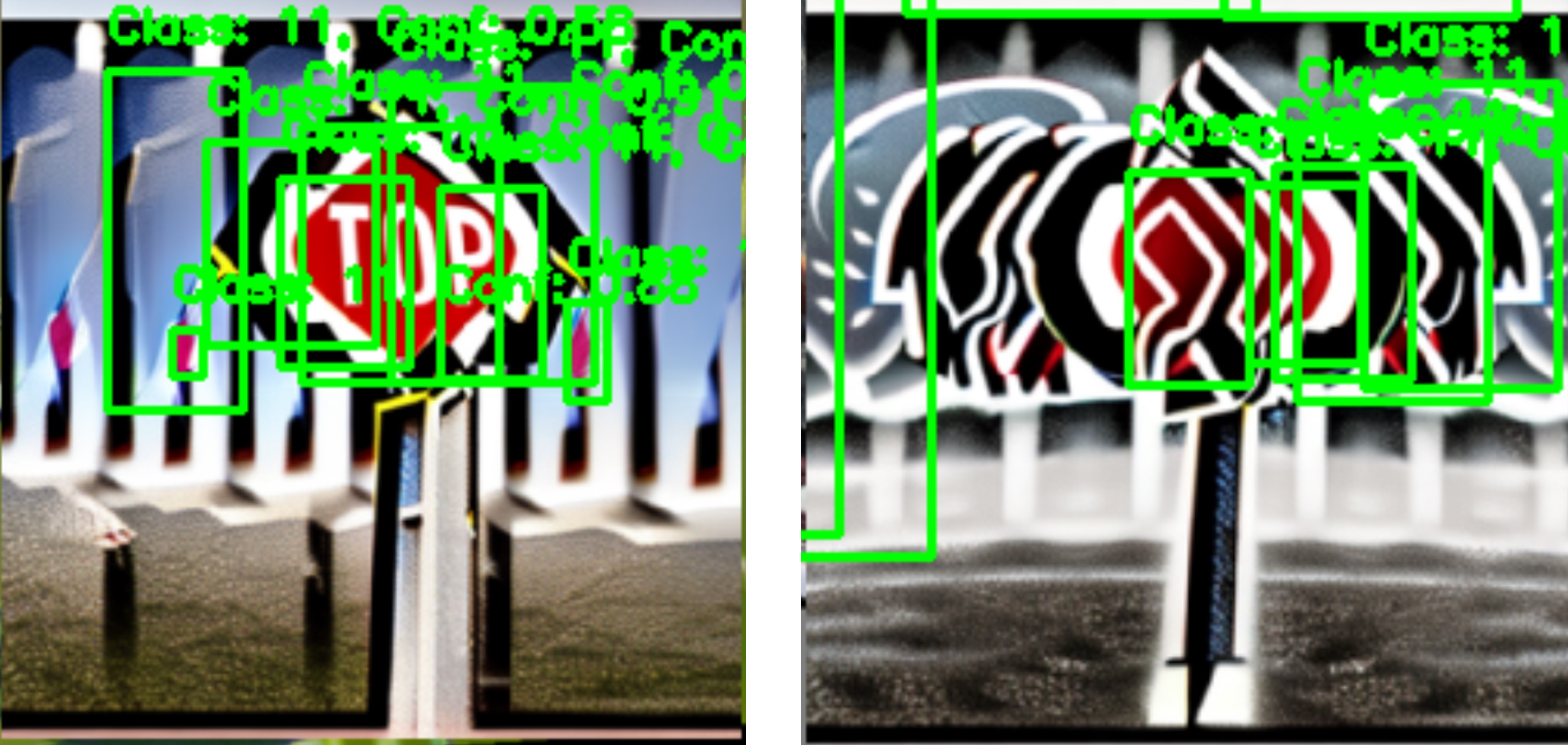}
\label{fig_3b}
}
\caption{Motivation examples of our crop mechanism and BBOX filter design: (a) crop mechanism to focus optimization on primary objects and (b) the necessity of a BBOX filter.}
\vspace{-0.3cm}
\label{fig:bad_case}

\end{figure}

In the initial stage, the latent variable $X_{t=T}$ is initialized as Gaussian noise, i.e. $X_T \sim\mathcal{N} (0, \textbf{I})$. For unconditional text embedding $\Phi_{t=T}$, it is initialized by passing a NULL text through the text encoder. Then, the image generated by diffusion model $X_0$ is input into the TSR. Previous diffusion-based attacks on classification and object detection~\cite{chen2024content,miao2024advlogo} optimize the latent variable $X_t$ by computing gradients over the entire $X_0$. However, this approach often results in the detection of multiple traffic signs in the background rather than on the primary object, i.e., C1 in~\S\ref{sec:challenge}, as illustrated in Fig.~\ref{fig:bad_case} (a). This deviation causes the generated image to stray from the intended prompt, undermining the original attack goal. 

To address this issue and prevent the optimization from focusing on non-primary objects, we introduce a cropping mechanism to achieve the optimization objective. This mechanism leverages the insight that, during the initial iterations, the TSR tends to identify the primary object as the target traffic sign~\cite{sato2024intriguing}. This mechanism directs the optimization process to concentrate on the main object: if the primary object can be detected by the TSR, it is cropped out for further iterative optimization; otherwise, the entire image is used. To further improve the attack robustness against different real-world perspectives, such as varying angles, we apply image transformations and embed the image into a real-world background to better simulate real-world conditions.

Moreover, we design novel loss functions: detection loss $L_d$ and similarity loss $L_s$, as introduced in~\S\ref{eq:score_loss}, leveraging the CLIP model~\cite{radford2021learning} to ensure that the generated images align well with both the prompt and the attack objective, which addresses C2 in~\ref{sec:challenge}. Additionally, we offer attack style customization from both image and prompt perspectives to improve our \system, which addresses C3 in~\ref{sec:challenge}.

\subsection{Loss Function Design}
To generate our attack, we design two loss functions: detection loss $L_d$ and similarity loss $L_s$.

{\bf Detection Loss.} The detection loss is for effective attack generation, causing the TSR system to misidentify an object as a specific target class, $T_\mathrm{class}$, by introducing adversarial properties into the images produced by the T2I diffusion model. The TSR without non-maximum suppression (NMS), $D$, takes an image $X$ as input and outputs proposal BBOXes $C_p$, which include the object location coordinates $C^\mathrm{coor}_p$, confidence scores for all classes $C^\mathrm{score}_p$, and the predicted classes of the proposal BBOXes $C^\mathrm{class}_p$:
\begin{center}
\label{eq:detect}
$ \{C^\mathrm{coor}_p, C^\mathrm{score}_p, C^\mathrm{class}_p\} = D(X)$
\end{center}
The input image $X$ of the TSR $D$ is defined as:
\begin{center}
$$ X=\left\{
\begin{array}{rcl}
\mathrm{emb}(b, \mathrm{Crop}(X_0,C_0^\mathrm{Coor}), \mathrm{Coor}_e), & {T_\mathrm{class} \in C_0^\mathrm{class}}\\
\mathrm{emb}(b, X_0, \mathrm{Coor}_e),\ \ \ \ \ \ \ \ \ \ \ \ \ \ \ \ \ \          & {T_\mathrm{class} \notin C_0^\mathrm{class}}
\end{array} \right. $$
\end{center}
where $\mathrm{emb}(b, x, \mathrm{Coor}_e)$ is an embedding function that integrates the image $x$ into the background image $b$ based on the randomly generated coordinate $\mathrm{Coor}_e$; and $C_0$ represents the BBOXes output by TSR with NMS $D^\star$ by inputting $X_0$.

To generate the attack effectively and efficiently, we filter the top $k$ BBOXes related to the target object to compute the detection loss $L_d$, similar to prior research~\cite{jia2022fooling}. Fig.~\ref{fig:bad_case} (b) illustrates the impact of not using a BBOX filter, which results in numerous detection outputs in the background. This makes attack less effective. Thus, the detection loss $L_d$ is defined as:
\begin{center}
\label{eq:score_loss}
${{L_{d} = {\frac{1}{|C_f|}} \sum\limits_{{c}\in C_f^\mathrm{score}}(\mathrm{ReLU}(T_\mathrm{score}-c[{T_\mathrm{class}}]))^2}}$\\
\end{center}
where $C_f$ represents the top $k$ BBOXes, defined as $C_f = C_p[\textbf{top}(\text{IOUs}, k)]$. The IOU is calculated as:
\begin{center}
$\text{IOUs}(C^\mathrm{coor}_p, \mathrm{Coor}_e) = \frac{\textbf{area}({C^\mathrm{coor}_p} \cap \mathrm{Coor}_e)}{\textbf{area}({C^\mathrm{coor}_p} \cup \mathrm{Coor}e)}$\\
\end{center}
where $T_\mathrm{score}$ denotes the confidence threshold set in the TSR.

{\bf Masked Prompt.} To prevent generated attacks from containing key features found in benign images, we propose a method that directs the T2I diffusion model to emphasize disrupting intrinsic features. For example, if the target class is a STOP sign, we do not want the attack to include the word ``STOP'', which can be easily noticed by humans. Instead, we aim to create images with semantically adversarial features, such as patterns resembling ``with a ghost on it''. This is achieved by designing a masked prompt $\rho_{m}$, which is formulated by removing the ``benign prompt'' from the original and retaining only the ``adversarial prompt''.

{\bf Similarity Loss.} To ensure that the optimization process emphasizes the masked prompt $\rho_m$, we utilize the CLIP model~\cite{radford2021learning}, which consists of both a text encoder $E_t$ and an image encoder $E_i$, to compute the cosine similarity between the input image $X$ and the masked prompt $\rho_{m}$. This similarity metric evaluates their semantic alignment, leveraging CLIP’s robust multimodal capabilities to accurately capture the relationship between image content and target text. The cosine similarity loss is defined as:
\begin{equation}
    L_s\left( \rho_{m}, X \right)=(1-\frac{{E_{i}}\left( X \right)\cdot {E_{t}}\left( \rho_{m} \right)}{\Vert {E_{i}}\left( X \right) \Vert_2 \cdot \Vert {E_{t}}\left( \rho_{m} \right) \Vert_2})
\end{equation}

Thus, we summarize the total loss as:
\begin{equation}
    L_{\theta}=L_s + \lambda \cdot L_d
\end{equation}
where $\lambda$ is a hyper-parameter to weight loss functions.

\begin{figure}[!t]
\centering
{\includegraphics[height=0.33\linewidth]
{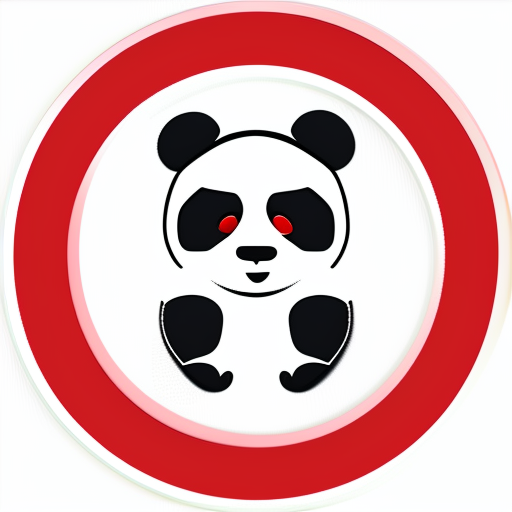}
}
{\includegraphics[height=0.33\linewidth]
{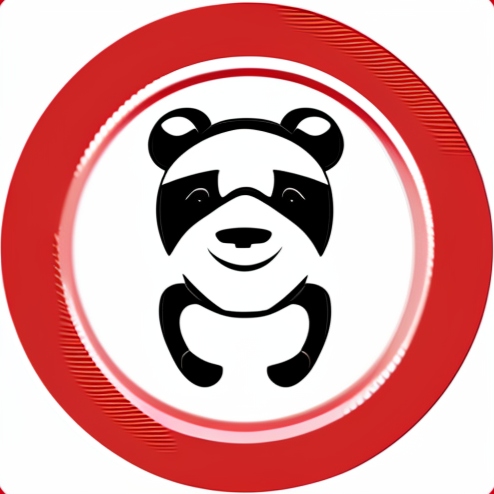}
}

\vspace{0.1cm}
\caption{Visualization examples of image-specified style customization. Left is the specified style image. Right is generated adversarial appearance attack.}
\vspace{-0.4cm}
\label{fig:style-customization}

\end{figure}

\subsection{Style Customization}
\label{sec:style}
As noted in C3 (\S\ref{sec:challenge}), T2I diffusion models struggle with out-of-domain objects due to limited prior knowledge. To address this, we introduce style customization techniques that inject semantic priors and guide the diffusion process. These methods enhance \system’s effectiveness on unfamiliar sign types and improve stealthiness by allowing control over shape and texture~\cite{zhao2019seeing, song2018physical}.

\textbf{Image-Specified Style Customization.} Attackers can generate attacks that match the style of a specified image, as illustrated in Fig.~\ref{fig:style-customization}. Specifically, we employ the NULL-text DDIM inversion method~\cite{mokady2023null} by using an image to initialize the latent variables $X_T$ instead of using random Gaussian noise. This approach allows the T2I diffusion model to search for adversarial attacks with semantics that preserve similarity to the reference image. Additionally, while prompts can be manually specified, they can also be generated automatically using image captioning models (e.g., BLIP v2~\cite{li2023blip}). 

\begin{figure}[!t]
\centering
\subfloat[w/o customization]
{\includegraphics[height=0.33\linewidth]
{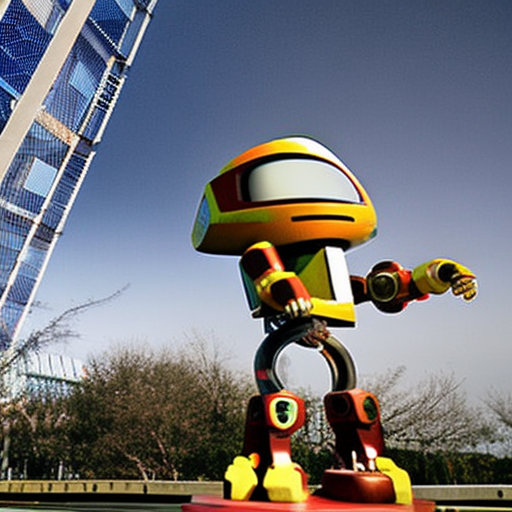}
}
\hspace{0.6cm}
\subfloat[with customization]
{\includegraphics[height=0.33\linewidth]
{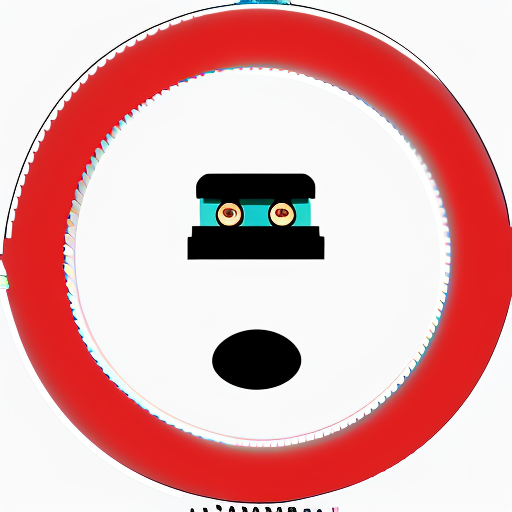}
}

\caption{Comparison of example generated with and without prompt-specified style customization. (a) without customization, the robot's leg joints are recognized as a speed limit sign. (b) with customization and the same prompt as (a).}

\label{fig:prompt-custom}

\end{figure}

\textbf{Prompt-Specified Style Customization.} In this method, attackers can use prompts to modify specific features on the benign image and generate customized attacks, e.g., replacing the text in a speed limit sign with a robot. Compared to non-customized attacks, attacks generated by this method are based on benign image controllable as shown in Fig.~\ref{fig:prompt-custom}. Specifically, we use a benign image as the style image, such as a standard speed limit sign indicating 5 km/h. Then, we initialize the latent variable $X_T$ using the benign image with the NULL-text DDIM inversion method. Finally, we replace the benign part of the prompt describing the benign image with the adversarial part desired by the attacker. We find that this method not only allows for style customization but also leverages the powerful generative capabilities of T2I diffusion models to produce natural and highly effective adversarial examples.

\subsection{Optimization Method}
To reduce computational costs and generate our attack more efficiently, we adopt gradient approximation for back-propagation, as prior works~\cite{miao2024advlogo, chen2024content}. Based on chain rule, the gradients $\nabla_{X_T}L_{\theta}$ and $\nabla_{\Phi_T}L_{\theta}$ can be obtained:
\begin{equation}
\begin{aligned}
\nabla_{X_T}L_{\theta}&=\frac{\partial L_{\theta}}{\partial X_0}\cdot\frac{\partial X_0}{\partial X_1}\cdot\frac{\partial X_1}{\partial X_2}\ldots\frac{\partial X_{T-1}}{\partial X_T} \\
\nabla_{\Phi_T}L_{\theta}&=\frac{\partial L_{\theta}}{\partial X_0}\cdot\frac{\partial X_0}{\partial X_1}\cdot\frac{\partial X_1}{\partial X_2}\ldots\frac{\partial X_{T-1}}{\partial \Phi_T}
\end{aligned}
\end{equation}
To approximate $\nabla_{X_T}L_{\theta}$, we leverage the following approximation function from DDIM~\cite{song2020denoising}:
\begin{equation}
\label{equ:ddim}
    X_t = \sqrt{\alpha_t}X_0 + \sqrt{1-\alpha_t}\varepsilon, \quad \varepsilon \sim \mathcal{N}(0, \mathbf{I}).
\end{equation}
where the $\alpha$ is variance schedule~\cite{song2020denoising} and the $\varepsilon$ is a noise variable. By rearranging Eq.~\eqref{equ:ddim}, we can express $X_0$ as $X_0 = \frac{1}{\sqrt{\alpha_t}}X_t - \sqrt{\frac{1 - \alpha_t}{\alpha_t}}\varepsilon$. Thereby, we obtain $\frac{\partial X_0}{\partial X_t} = \frac{1}{\sqrt{\alpha_t}}$. Thus, the gradient of $X_T$ can be approximated as: 
\begin{equation}
\begin{aligned}
    \label{eq:lossx}
    \nabla_{X_T}L_{\theta}&=\frac{1}{\sqrt{\alpha_T}}\cdot\nabla_{X_0}L_{\theta}
\end{aligned}
\end{equation}

To calculate the gradient over $\Phi_T$, using the approximation in Eq.~\eqref{equ:ddim} alone is insufficient due to the absence of a relationship between $X$ and $\Phi$. Therefore, we utilize an additional equation provided in DDIM~\cite{song2020denoising}.
\begin{align}
\label{equ:iteration}
    \ X_{t-1}(X_{t},t, \eta,\Phi_t)=\sqrt{\frac{\alpha_{t-1}}{\alpha_t}}X_t + \ \ \ \ \ \ \ \ \ \   & \notag\\ 
    \sqrt{\alpha_{t-1}}\left(\sqrt{\frac{1}{\alpha_{t-1}}-1}-\sqrt{\frac{1}{\alpha_t}-1} 
    \right)& \cdot \tilde{\epsilon}_\theta(X_t,t,\eta,\Phi_t)
\end{align}
where $\tilde{\epsilon}_\theta$ is optimal model in DDIM.

Thus the gradient of $\nabla_{\Phi_T}L_{\theta}$ can be approximated as:
\begin{equation}
\begin{aligned}
    \label{eq:lossphi}
    \nabla_{\Phi_T}L_{\theta}&=\nabla_{X_0}L_{\theta} \cdot \frac{\partial X_0}{\partial X_{T-1}} \cdot \frac{\partial {X_{T-1}}}{\partial \Phi_T}\\
    &=\nabla_{X_0}L_{\theta} \cdot \frac{1}{\sqrt{\alpha_{T - 1}}} \cdot \sqrt{\alpha_{T-1}} \cdot \\ 
    &\left(\sqrt{\frac{1}{\alpha_{T-1}}-1}-\sqrt{\frac{1}{\alpha_T}-1} 
    \right) \cdot \frac{\partial \tilde{\epsilon}_\theta}{\partial \Phi_T} \\
    &=\nabla_{X_0}L_{\theta} \cdot \left(\sqrt{\frac{1}{\alpha_{T-1}}-1}-\sqrt{\frac{1}{\alpha_T}-1} 
    \right) \cdot \frac{\partial \tilde{\epsilon}_\theta}{\partial \Phi_T}
\end{aligned}
\end{equation}

With Eq.~\eqref{eq:lossx} and Eq.~\eqref{eq:lossphi}, we could further optimize the $X_T$ and $\Phi_T$ as $X_t  \leftarrow X_t - \beta \cdot \nabla_{X_T}L_{\theta}$ and $\Phi_T  \leftarrow \Phi_T - \gamma \cdot \nabla_{\Phi_T}L_{\theta}$, where $\beta$ and $\gamma$ are two hyper-parameters.

\section{Evaluation}
\label{sec:eval}

\begin{figure*}[!t]
\centering
\subfloat[Sunny, STOP sign]
{\includegraphics[width=0.32\linewidth]
{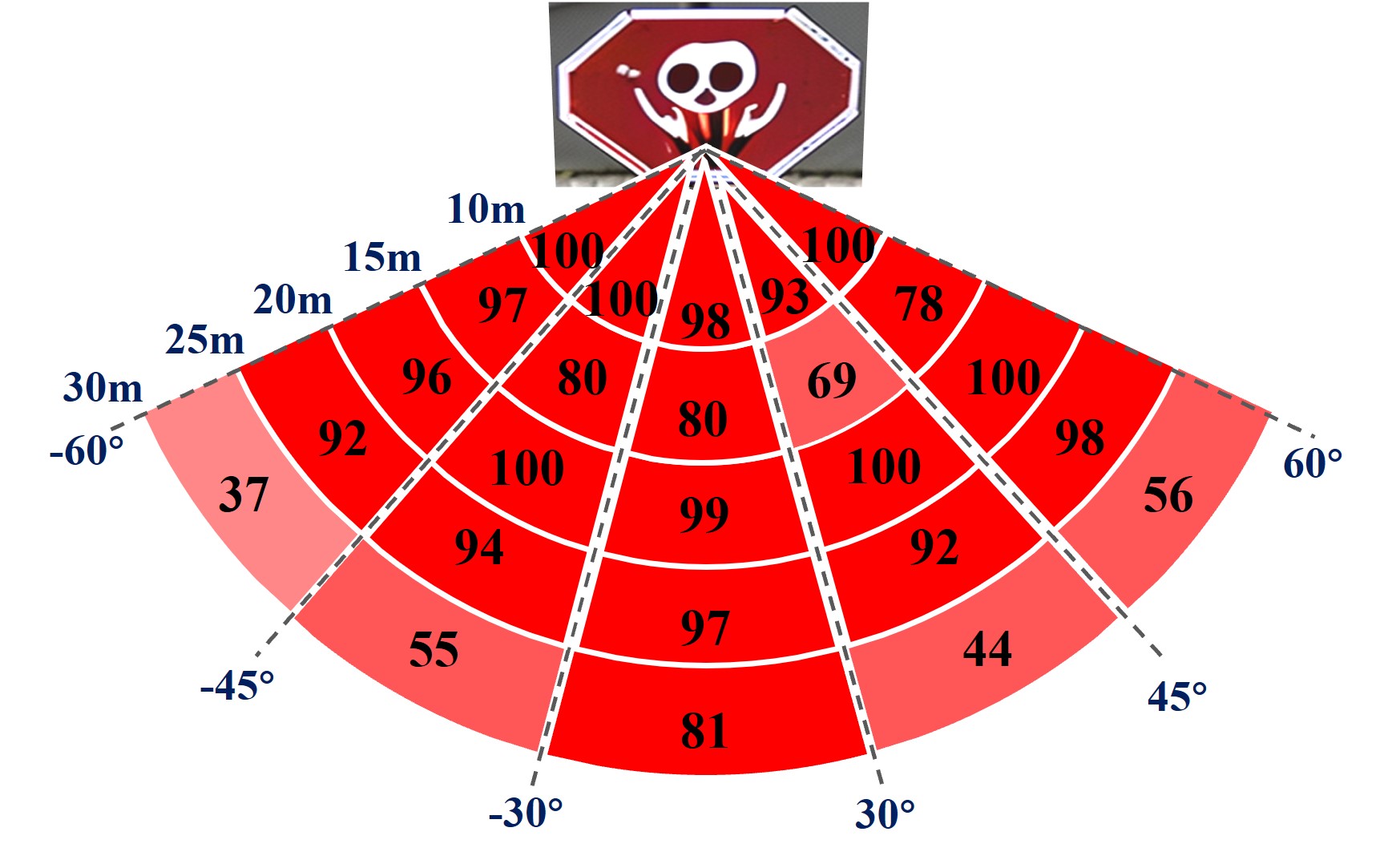}
\label{fig_phy1}
}
\subfloat[Cloudy, STOP sign]
{\includegraphics[width=0.32\linewidth]
{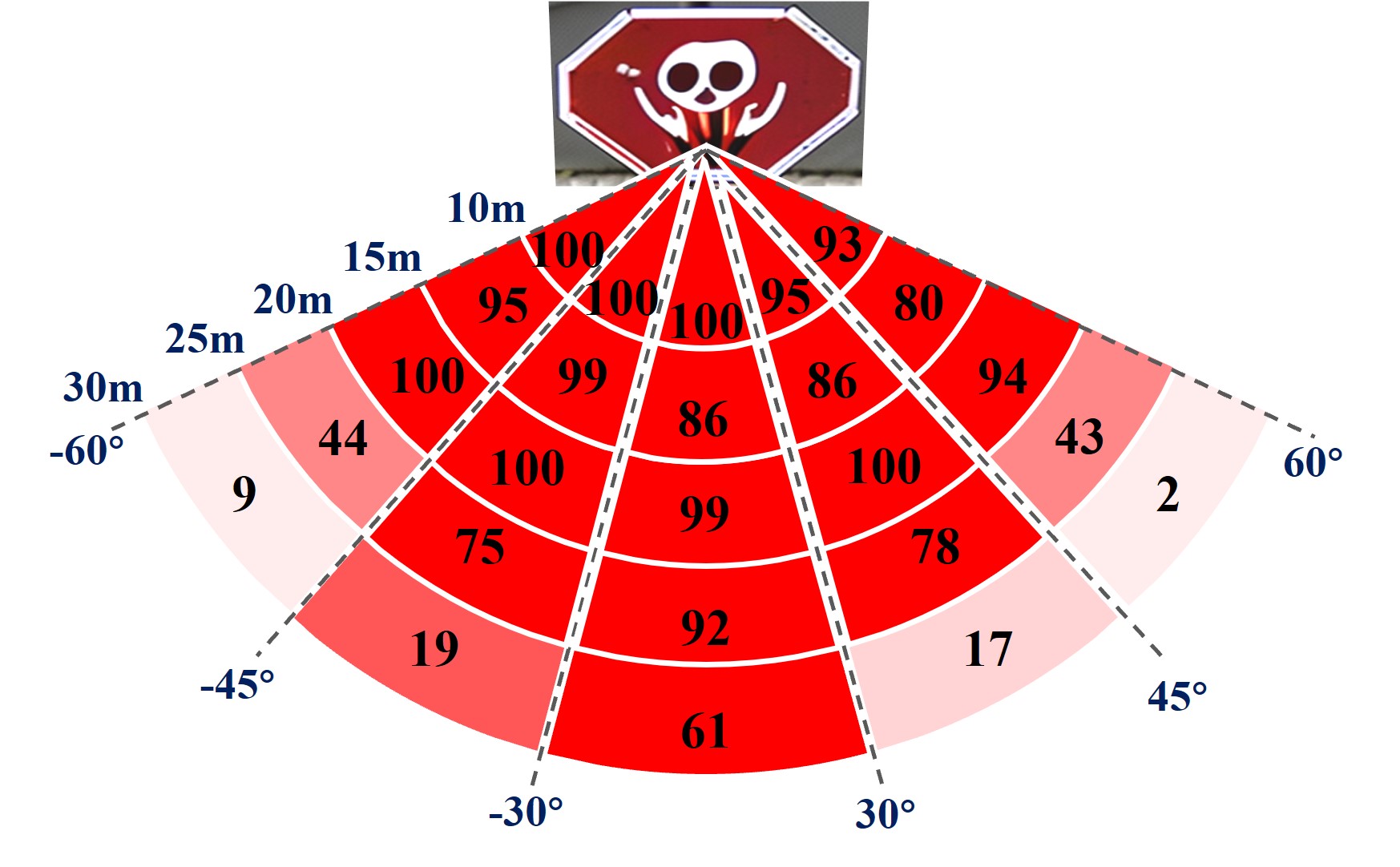}
\label{fig_phy2}
}
\subfloat[Night, STOP sign]
{\includegraphics[width=0.32\linewidth]
{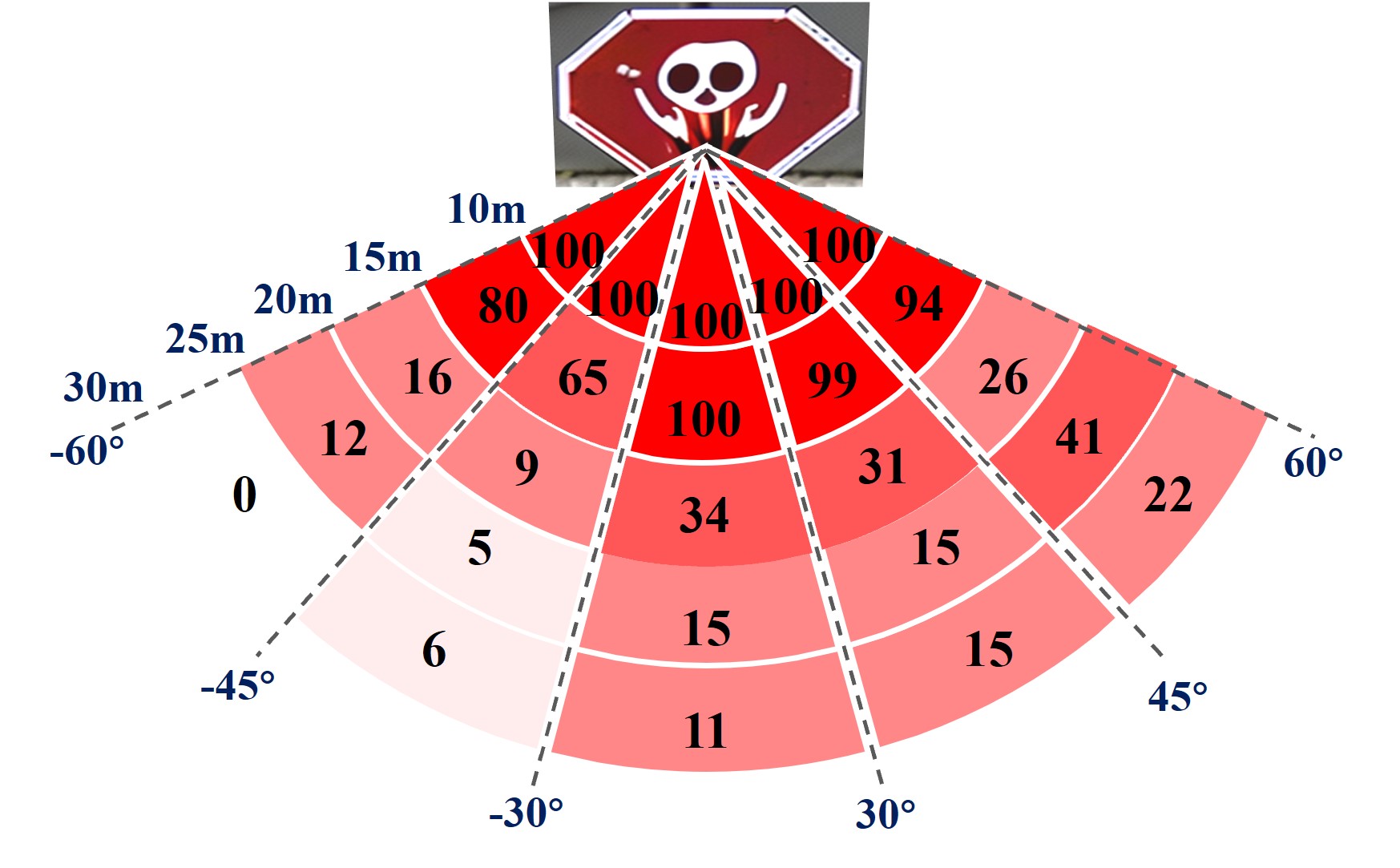}
\label{fig_phy3}
}
\hfill
\subfloat[Sunny, Speed limit]
{\includegraphics[width=0.32\linewidth]
{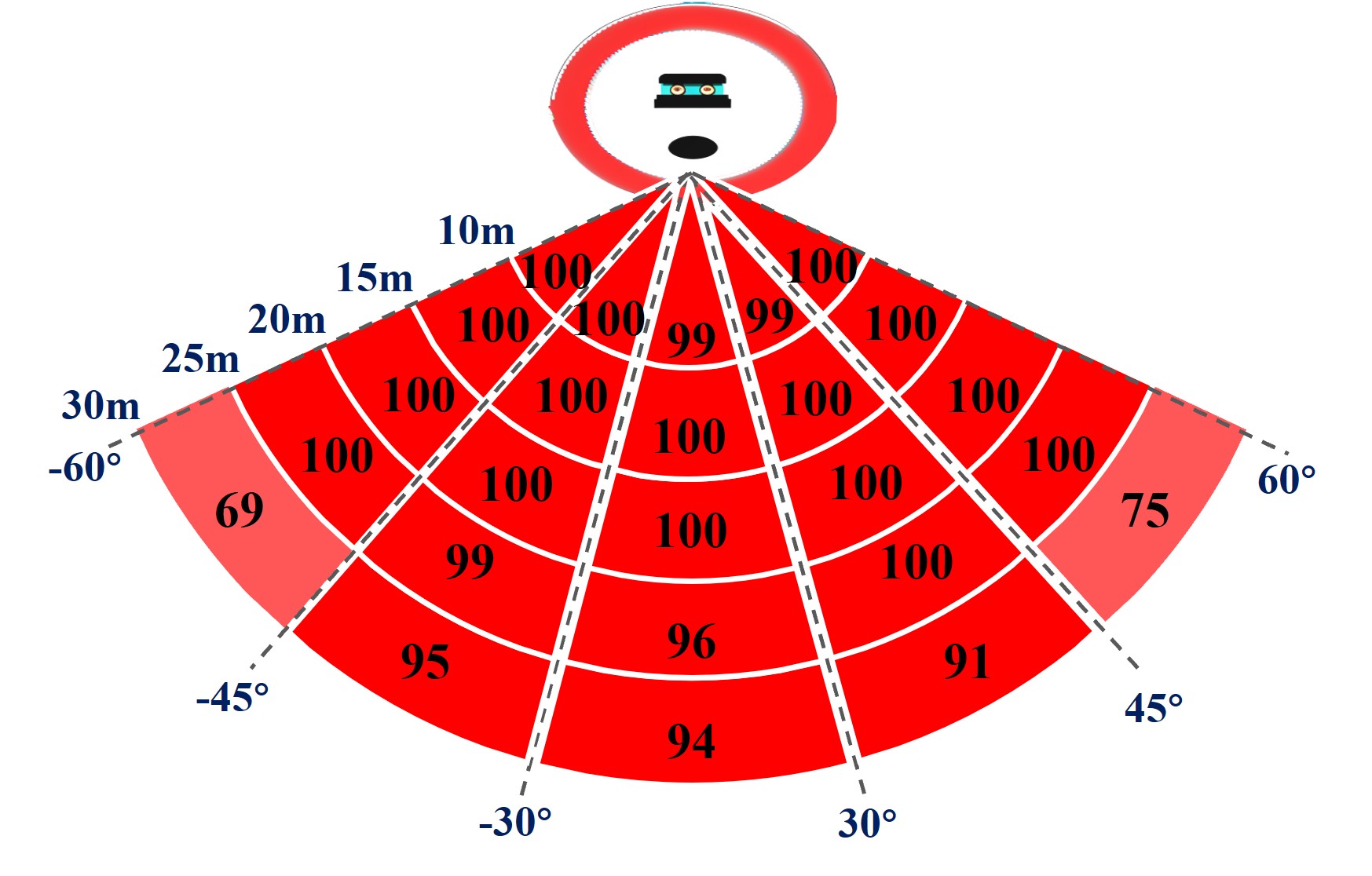}
\label{fig_phy4}
}
\subfloat[Cloudy, Speed limit]
{\includegraphics[width=0.32\linewidth]
{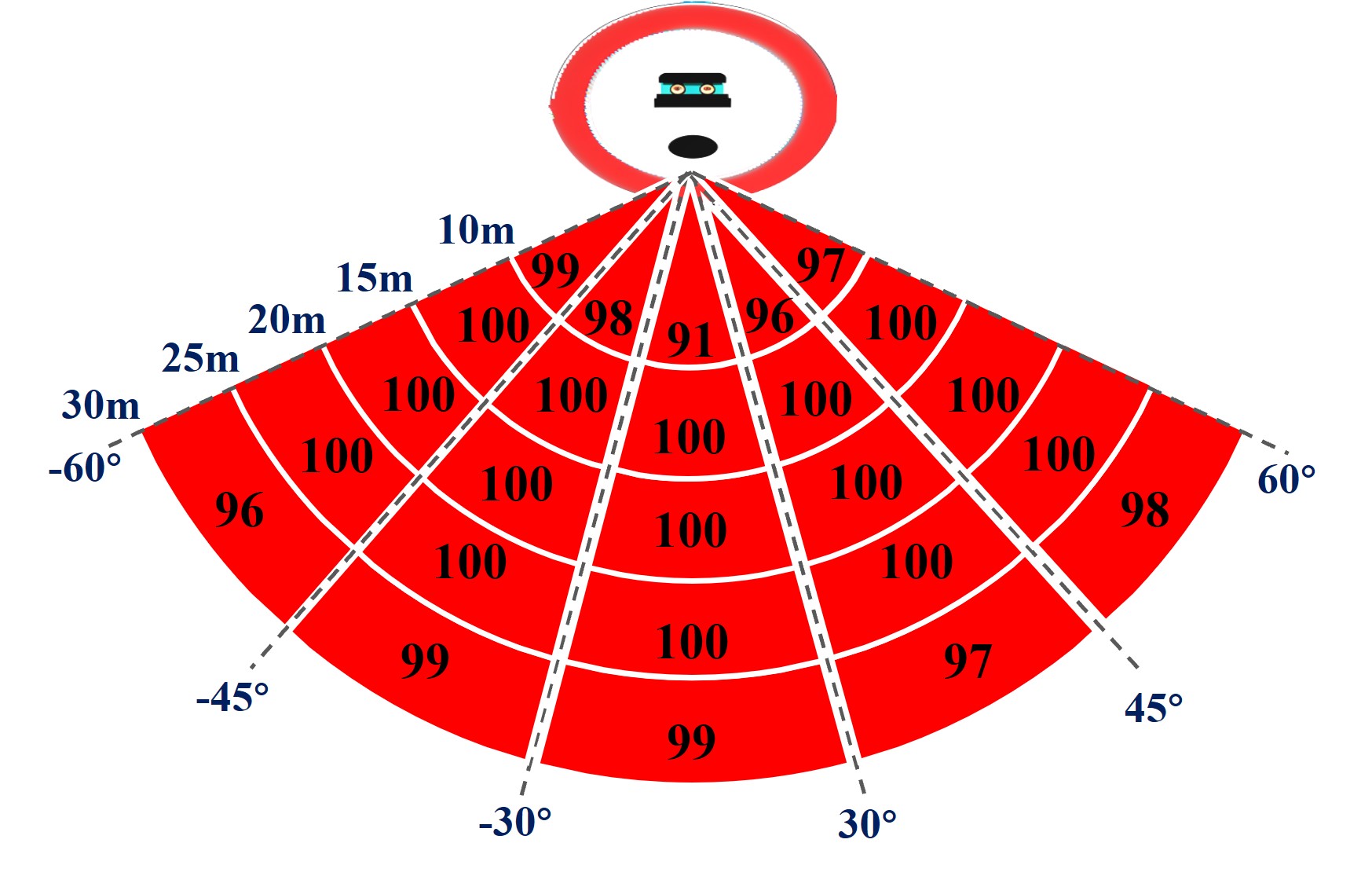}
\label{fig_phy5}
}
\subfloat[Night, Speed limit]
{\includegraphics[width=0.32\linewidth]
{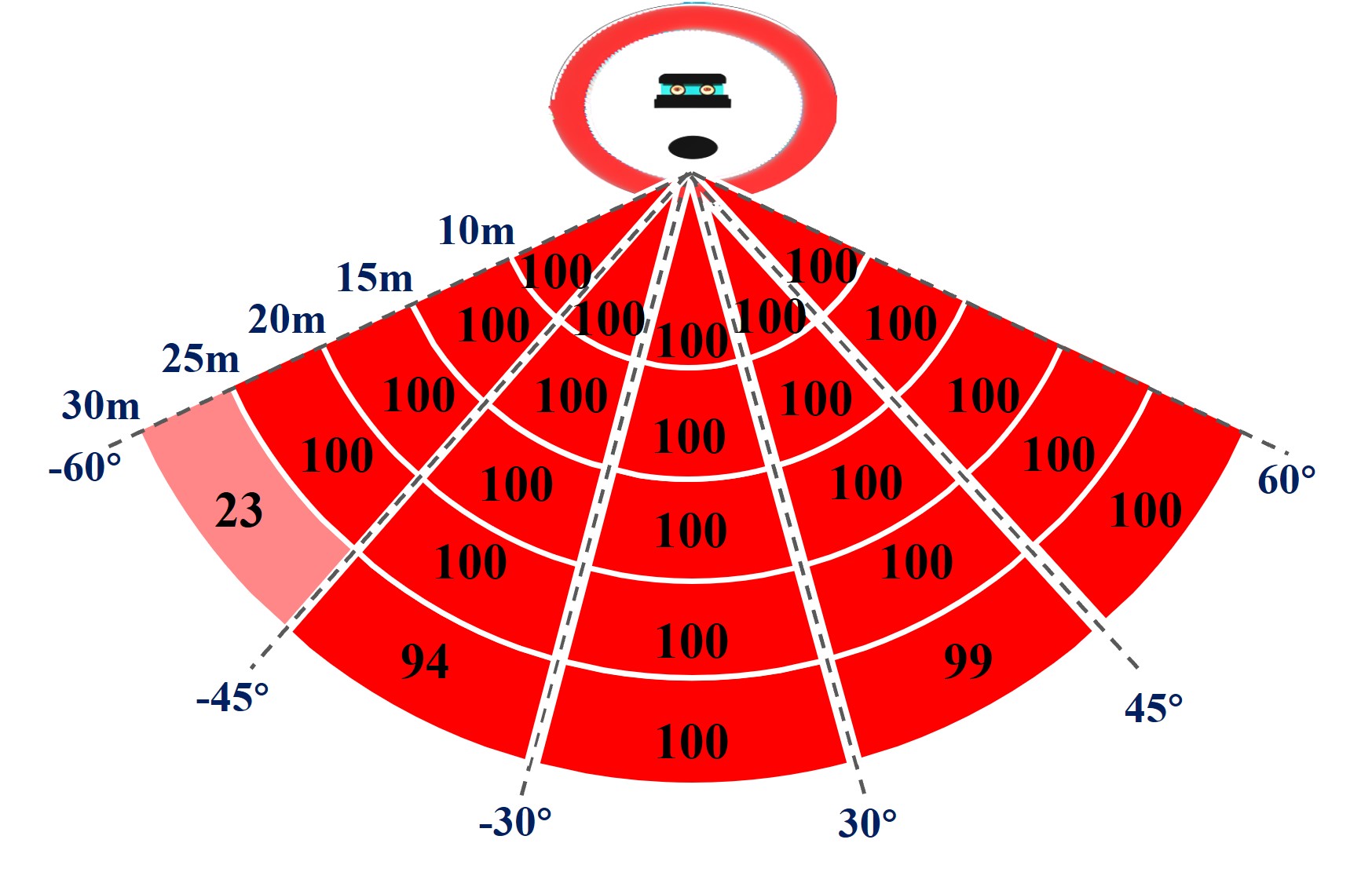}
\label{fig_phy6}
}
\caption{Attack success rate for both STOP sign and speed limit sign in physical world at different angles, distances, and light conditions. The depth of red represents the value of ASR, whereas the darker color indicates a higher attack success rate.}
\vspace{-0.2cm}
\label{fig:phy}

\end{figure*}

\subsection{Experimental Setup}
\label{sec:exp_setup}
{\bf Target Traffic Sign Selection.} To evaluate the generalization capability of \system, we select 6 traffic signs covering different categories: prohibitory signs (STOP, No Entry), warning signs (Children Crossing), and speed limits (5, 40, 60 km/h). Among these, we primarily focus on the STOP sign and the 5 km/h speed limit sign as our main targets because (1) STOP and 5 km/h speed limit signs are popular targets in prior work for demonstrating adversarial attack effects in the physical world~\cite{wang2025revisiting, Wang_2023_ICCV, eykholt2018physical, sato2024intriguing, sato2024invisible, zhao2019seeing, jia2022fooling}, and (2) both are highly safety-critical. For example, a falsely appearing 5 km/h speed limit (an unusually low speed) could trigger rear-end collisions when the ego vehicle suddenly brakes. Detailed evaluation results for other sign types are presented in Appendix.

{\bf Dataset and Models.} Two datasets are selected for TSR tasks: COCO~\cite{lin2014microsoft}, for STOP sign detection, and TT100k~\cite{Zhe_2016_CVPR}, for speed limit sign detection. We use YOLO v5 as our object detector introduced in~\S\ref{sec:method}, given its success in demonstrating transfer attacks on commercial vehicle models~\cite{jia2022fooling, wang2025revisiting}. For STOP sign detection, we utilize the pre-trained YOLO v5 model from the official website~\cite{glenn_jocher_2020_4154370}. For speed limit sign detection, we retrain YOLO v5 on the TT100k dataset. Consistent with the settings in previous work~\cite{jia2022fooling}, each 2048×2048 image was divided into 16 smaller 640×640 images, and we focused on the top 50 traffic sign categories by data volume. As a result, YOLO v5 achieved a mean Average Precision (mAP) of 79.5\%. Stable Diffusion 2.1~\cite{rombach2022high} is employed as the base model in the DDIM denoising process, detailed in~\S\ref{sec:method}. We further investigate the transferability of our attack across different models, such as Faster-RCNN, YOLO v11, etc, with the default official settings. Additionally, we retrain Faster R-CNN for speed limit sign on the TT100k dataset for transferability evaluation, achieving an mAP of 74.7\%. 

{\bf Attack Printing.} To evaluate the attack in physical world, we print the generated appearance attack as a 60 cm × 60 cm poster to represent our \system, which is the same representative setup as prior research~\cite{zhao2019seeing, Wang_2023_ICCV, wang2025revisiting}.

{\bf Evaluation Metrics.}
We evaluate the attack effectiveness of our \system using the Attack Success Rate (ASR), where a successful attack is defined as the TSR recognizing the specified traffic sign given videos. The ASR is calculated as:
$\mathrm{ASR} = \frac{N_\mathrm{succ}}{N_\mathrm{all}}$, 
where $N_\mathrm{succ}$ is the number of frames for successful attack, and  $N_\mathrm{all}$ is the total number of frames.

{\bf Hardware.} We generate all attacks and conduct all experiments on a single NVIDIA Tesla A100 40GB GPU with CUDA Version 12.5 and Driver Version 555.58.02, paired with an Intel(R) Xeon(R) Gold 6226R CPU @ 2.90 GHz. Our implementation is in Python 3.8.20 using PyTorch 1.13.1.

\begin{table*}[t!]
\small
\tabcolsep 0.05in
    \caption{Physical-world attack transferability evaluation over 6 target detectors including both one-stage and two-stage ones: YOLO v3~\cite{redmon2018yolov3}, YOLO v4~\cite{bochkovskiy2020yolov4}, YOLO v11~\cite{yolo11_ultralytics}, Faster-RCNN~\cite{ren2015faster}, Mask-RCNN (MR)~\cite{he2017mask}, and DETR~\cite{carion2020end}. For the speed limit case, all detectors are trained on the TT100K dataset~\cite{Zhe_2016_CVPR}. Note that YOLO v5~\cite{Jocher2022yolov5} is source model for comparison.}
    \centering
    \begin{tabular}{cccccccccc}

    \toprule
    & \multicolumn{7}{c}{STOP Sign} & \multicolumn{2}{c}{Speed Limit (5 km/h)}\\
    \cmidrule(lr){2-8}
    \cmidrule(lr){9-10}
    Distance & YOLO v3 & YOLO v4 & YOLO v5 & YOLO v11 & Faster R-CNN & Mask R-CNN & DETR & YOLO v5 & Faster R-CNN \\
    
    \midrule
    $0 \sim 5m$ & 100\% & 100\% & 100\% & 100\% & 100\% & 100\% & 100\% & 99\% & 53\%\\
    $5m \sim 10m$ & 100\% & 100\% & 96\% & 100\% & 100\% & 100\% &100\% & 100\% & 100\%\\
    $10m \sim 15m$ & 99\% & 93\% & 80\% & 95\% & 100\% & 100\% &99\% & 100\% & 100\%\\
    $15m \sim 20m$ & 82\% & 44\% & 99\% & 87\% & 100\% & 100\% &96\% & 100\% & 94\%\\
    $20m \sim 25m$ & 85\% & 32\% & 97\% & 41\% & 100\% & 98\% &100\% & 96\% & 85\%\\
    $25m \sim 30m$ & 98\% & 17\% & 81\% & 62\% & 100\% & 91\% &97\% & 94\% & 70\%\\
    
\midrule
\midrule
    Average & 94\% & 64\% & 92\% & 81\% & 100\% & 98\% &99\% & 98\% & 84\%\\

    \bottomrule
         
    \end{tabular}
    \vspace{-0.2cm}

    \label{tab:trans_phy}
\end{table*}

\subsection{Attack Effectiveness in Physical World}
\label{sec:attack_effectiveness}
In this section, we evaluate the attack effectiveness of \system in the physical-world environment.

{\bf Methodology and Setup.} We evaluate our \system in an outdoor physical-world environment with three different light conditions: sunny ($\approx 18,000\ lux$), cloudy ($\approx 7,000\ lux$), and night ($\approx 80\ lux$). The camera used for recording videos has the same configurations—such as focal length and resolution—as the cameras used in Baidu Apollo~\cite{apollo}, an industry-grade AD system. We record videos from far to near ($0\sim30m$) at different angles, respectively, which is a similar setup as prior research on TSR system security~\cite{zhao2019seeing, jia2022fooling}.

{\bf Results.} Fig.~\ref{fig:phy} shows the ASR of \system under varying distances, angles, light conditions, and sign types, with an 83.3\% overall ASR. The STOP sign appearance attack performs well on sunny and cloudy days, with ASRs of 85.4\% and 74.7\%, respectively, but shows reduced effectiveness at night (47.8\%), especially at larger distances. This decrease might be due to the low visibility at night on appearance attacks, a phenomenon also observed in prior research~\cite{jia2022fooling}. Due to the spatial memorization effect commonly observed in commercial TSR systems observed in prior research~\cite{wang2025revisiting}, the ASR illustrated in Fig.~\ref{fig:phy} can achieve 100\% attack effectiveness at the TSR system level, even during nighttime. For the speed limit sign attack effectiveness in Fig.~\ref{fig:phy}, \system demonstrates very high attack effectiveness across all light conditions: 97.7\% on sunny days, 98.8\% on cloudy days, and 96.6\% at night. Such high attack effectiveness suggests the promising potential for positive results in commercial testing in~\S\ref{sec:commercial}. Visualization examples of real-world attack video frames—captured under varying distances, viewing angles, light conditions, and sign types—are provided in Appendix.

\begin{table*}[t]
\tabcolsep 0.07in
\centering
\small
    \caption{Baseline comparison on STOP sign case between our attack \system and baselines: NDD~\cite{sato2024intriguing}, $\text{NDD}^{\ast}$~\cite{sato2024intriguing}, and SIB~\cite{zhao2019seeing}. NDD denotes generated objects without text `STOP', $\text{NDD}^{\ast}$ denotes generated objects with text `STOP'. AASR denotes the average ASR across these eight object detectors. The NDD, $\text{NDD}^{\ast}$, and \system are generated using YOLO v5 as the surrogate detector, whereas SIB is generated using YOLO v3. The surrogate detectors are selected based on the setup in the original paper. All attacks are further evaluated for their transferability on the remaining seven TSR object detectors.}

    \begin{tabular}{cccccccccc}

    \toprule
     &  \multicolumn{8}{c}{TSR Object Detectors (STOP sign)}   \\
     \cmidrule(lr){2-9}
   \multirow{-2}{*}{\shortstack{Attack Method}} &  YOLO v3 & YOLO v4 & YOLO v5 & YOLO v11 & Faster R-CNN & Mask R-CNN & DETR & SSD~\cite{liu2016ssd}& \multirow{-2}{*}{AASR$\uparrow$}\\

   \midrule
   NDD~\cite{sato2024intriguing}& 1.8\%& 1.6\%& 8.3\%& 11.9\% & 36.5\% & 44.4\%& 11.9\%& 6.0\% & 15.3\%\\
  $\text{NDD}^{\ast}$~\cite{sato2024intriguing}& 31.0\%& 28.2\%& 39.9\%& 44.4\% & 71.7\% & 78.2\%& \textbf{45.0\%}& 28.1\% & 45.8\%\\
   SIB~\cite{zhao2019seeing} & \textbf{98.6\%}&0.9\% & 0.5\%&  3.0\%&  20.4\% & 6.7\% & 4.3\% & 1.8\% & 17.0\%\\
   
   \midrule
   \system & 92.8\% & \textbf{78.6\%}& \textbf{82.2\%}& \textbf{84.4\%} & \textbf{99.9\%} & \textbf{100\%} & 42.0\% & \textbf{81.8\%} & \textbf{82.7\%}\\
    
    \bottomrule
         
    \end{tabular}

    \label{tab:trans}
    \vspace{-0.2cm}
\end{table*}

{\bf Physical-World Attack Transferability.} We further evaluate the physical-world attack transferability over different TSR models
including both types of object detection introduced in~\S\ref{sec:TSR_system}: YOLO v3~\cite{redmon2018yolov3}, YOLO v4~\cite{bochkovskiy2020yolov4}, YOLO v11~\cite{yolo11_ultralytics}, Faster R-CNN~\cite{ren2015faster}, Mask R-CNN~\cite{he2017mask}, and DETR~\cite{carion2020end}. 
STOP sign and 5 km/h speed limit signs are evaluated (\S\ref{sec:exp_setup}).
The experiments are conducted on a sunny day at 0°. 

The results of physical-world attack transferability are summarized in Table~\ref{tab:trans_phy}. For STOP sign case, \system achieves an overall average ASR of 90\% for attack transferability. Our newly generated attack demonstrates even better transferability to two-stage object detectors, such as Faster R-CNN and Mast R-CNN, with an average 95\% ASR. Even in the worst-case scenario, where the attack is transferred to YOLO v4, \system still achieves a 64\% ASR across all distances. These results may suggest that YOLO v4 has limited performance in detecting small objects. 
Furthermore, unlike prior pixel-level methods~\cite{zhao2019seeing,jia2022fooling} that overfit to their surrogate model detailed in~\S\ref{sec:Baseline}, \system does not exhibit disproportionately high performance on the surrogate model YOLO v5, indicating that it overcomes this limitation. For the speed limit case, we evaluate the transferability attack success rates on Faster R-CNN, achieving 84\% ASR. Detailed evaluation results for the speed limit sign case—covering variations in distance, viewing angles, and light conditions—are presented in Appendix.
These findings further indicate the high black-box potential of our attack, with promising implications for attacks on the commercial TSR system in~\S\ref{sec:commercial}.

\subsection{Baseline Comparison}
\label{sec:Baseline}
{\bf Baseline Attack Selection.} We compare \system with baseline attacks in this section, where two representative existing appearance attacks on TSR systems are selected: NDD~\cite{sato2024intriguing} and SIB~\cite{zhao2019seeing}. Since both of these attacks target STOP signs, we compare our \system with them specifically on STOP sign detection. For the NDD attack, we use the same prompt as our \system to ensure a fair comparison. Additionally, the NDD attacks can be classified into two versions: one without the `STOP' text on the object, denoted as NDD, and another with the `STOP' text, denoted as NDD*, for comparison.

{\bf Methodology and Setup.} We perform the baseline comparison with digital simulation, which is the same as prior research~\cite{cao2021invisible, zhao2019seeing}. The digital simulation applies random resizing and rotations, overlaying the generated appearance attacks onto various backgrounds, and performing a total of 1,000 different cases for evaluation. We ensure that image transformations are consistent across both \system and baseline attacks to ensure fairness. Specifically, we apply random resizing with scale factors between 0.4 and 1.0 (original size: 150×150) and rotations within a range of [-15°, 15°]. The transformed attacks are then placed on various real-world background images without STOP sign object from the COCO dataset, with each transformation applied simultaneously and randomized using a uniform distribution. In the case of SIB, we use the appearance attacks provided by the authors. The NDD, $\text{NDD}^{\ast}$, and \system are generated using YOLO v5 as the surrogate detector, whereas SIB is generated using YOLO v3. All attack methods are further evaluated for their transferability on the remaining seven TSR object detectors listed in Table~\ref{tab:trans_phy}. The surrogate detectors are selected based on the evaluation setup in the original paper. The average attack success rate (AASR) is calculated as the average of ASR across these eight models.

{\bf Results.} The results are summarized in Table~\ref{tab:trans}. As shown, \system significantly outperforms all baseline attacks in terms of AASR, achieving 82.7\%. This is approximately 5.4× higher than NDD (15.3\%), 1.8× higher than $\text{NDD}^{\ast}$ (45.8\%), and 4.9× higher than SIB (17.0\%). SIB shows strong performance on its surrogate model YOLO v3 (98.6\%), but fails to generalize to other detectors (mostly smaller than 5\%), indicating severe overfitting effects on its surrogate model. This highlights a core limitation of pixel-level methods, which often lack cross-model transferability. This limits its practical threat potential in real-world scenarios, especially for commercial TSR systems deployed on production AD vehicles, which are generally black-box to attackers. NDD and NDD* exhibit moderate transferability but poor effectiveness, with NDD* benefiting from the inclusion of the `STOP' text. This is due to a lack of detailed guidance for attack generation. In contrast, \system achieves consistently high ASRs across both one-stage and two-stage detectors in both white-box and black-box attack transferability settings. For instance, \system achieves 100\% ASR on Mask R-CNN and 99.9\% on Faster R-CNN, demonstrating strong black-box transferability and robust attack potential.

\begin{figure*}[!t]
\centering

{\includegraphics[width=0.95\linewidth]
{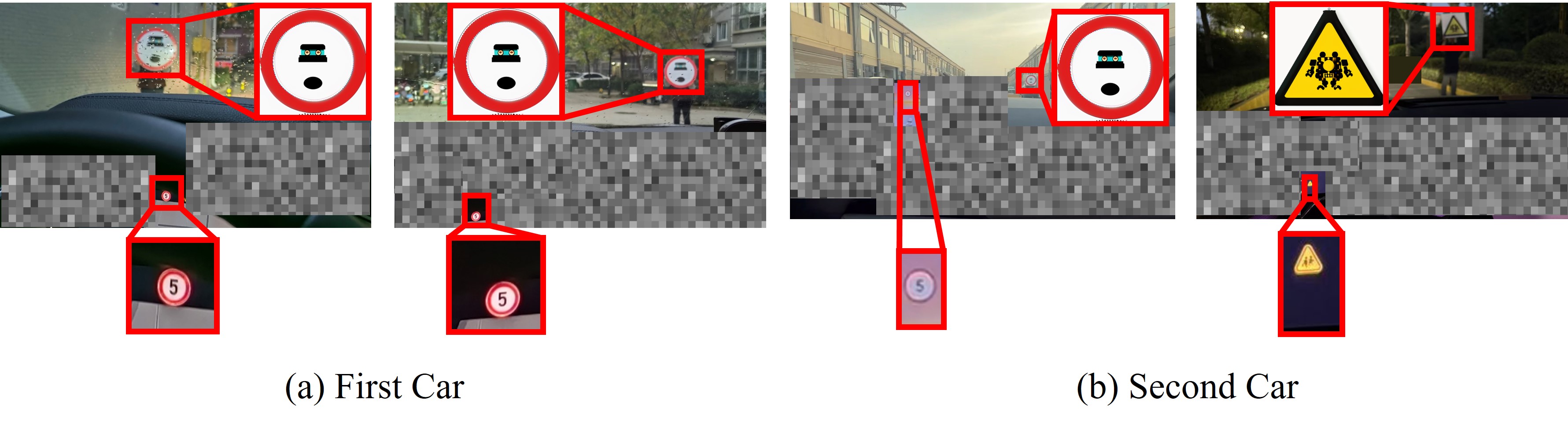}
}

\caption{Visualization of our attack \system effectiveness in commercial TSR systems from leading production vehicle brands. Detection results are shown on the dashboard or central display screen. Potential vehicle brand identifiers are hidden due to responsible vulnerability disclosure needs.}
\vspace{-0.3cm}
\label{fig:commercial}

\end{figure*}

\subsection{Attack Effect on Commercial TSR System}
\label{sec:commercial}
{\bf Methodology and Setup.} We evaluate the effectiveness of our proposed \system on two commercial TSR systems from leading production vehicle brands.
As one of tested vehicles do not support STOP sign detection, our evaluation mainly focuses on the speed limit sign appearance attack. To further assess the generalizability of our approach across different sign types, we additionally conduct attacks targeting the ``Children Crossing'' sign on another tested vehicles. We conduct experiments under real-world driving conditions on a sunny afternoon, testing sign placements both directly in front of the vehicle and off to the side. Each vehicle is driven past the sign ten times. An experimenter seated in the rear passenger seat observes the dashboard and central display to determine whether the vehicle recognizes the signs. An attack is considered successful if this recognition occurs. To mitigate spatial memorization effects in commercial TSR systems~\cite{wang2025revisiting}, we follow the recommended protocol from prior work~\cite{wang2025revisiting} by relocating the vehicle away from the test site after each trial and ensuring the system’s memory is cleared before next run.

{\bf Results and Visualization.} As shown in Fig.~\ref{fig:commercial}, after the vehicle passes by our adversarial attack, the 5 km/h speed limit sign or ``Children Crossing'' sign appear on the vehicle’s dashboard or central display screen, indicating that \system successfully fools the TSR systems of these commercial vehicles. \system achieves an overall 97\% attack success rate across ten runs for each scenario, demonstrating the high effectiveness of the commercial TSR system.

\subsection{Attack Qualification and Stealthiness}
\label{sec:user-study}

\begin{table}[t]
\centering
\small
\tabcolsep 0.1in
\renewcommand{\arraystretch}{1.25}
\caption{Attack qualification and stealthiness evaluation with user study. Qualification: the proportion of users who identify the image as the target class (e.g., STOP sign). Stealthiness: the proportion of users who consider the image likely to appear in daily life.}
\label{tab:user-study}
\begin{tabular}{|c|c|c|c|c|}
\hline
 \multicolumn{5}{|c|}{\textbf{Qualification} $\downarrow$} \\
\cline{1-5}
Benign & NDD~\cite{sato2024intriguing} & NDD*~\cite{sato2024intriguing} & SIB~\cite{zhao2019seeing} & \textbf{\system}  \\
\hline
\cellcolor{red!70} 90\% & \cellcolor{red!20} 26\% & \cellcolor{red!45} 68\% & \cellcolor{red!10} 21\%& \cellcolor{red!10} 21\%  \\
\hline
\multicolumn{5}{|c|}{\textbf{Stealthiness} $\uparrow$} \\
\cline{1-5}
FTE~\cite{jia2022fooling} & NDD~\cite{sato2024intriguing} &NDD*~\cite{sato2024intriguing} & SIB~\cite{zhao2019seeing}  & \textbf{\system} \\
\hline
\cellcolor{green!10} 14\% & \cellcolor{green!15} 18\% & \cellcolor{green!55} 63\% & \cellcolor{green!10} 12\%& \cellcolor{green!50} 41\%  \\
\hline
\end{tabular}
\vspace{-0.3cm}
\end{table}

{\bf Methodology and Setup.} To qualify as an appearance attack, two criteria must be met: (1) the generated object must not be perceived by human observers as the target sign, and (2) the attack must be stealthy: i.e., it should appear realistic and inconspicuous in everyday settings. To evaluate these criteria, we conduct a user study. This study underwent Institutional Review Board (IRB) review and, consistent with prior work~\cite{sato2021dirty, cao2021invisible}, was classified as IRB Exempt.

We recruited 121 human participants to evaluate both the qualification and stealthiness of various images, including benign samples, those generated by baseline attacks, and those produced by \system. To ensure fair and consistent comparison with prior work, our study adopts the core question used in NDD~\cite{sato2024intriguing}: ``Do you think there is a real [Sign Type] in this image?''—referred to as the qualification question. This question is well-suited for assessing qualification, as adversarial attacks are fundamentally characterized by a discrepancy between human-labeled ground truth and machine perception. For instance, a standard benign STOP sign would not qualify as an adversarial attack, as it is correctly perceived by both humans and models.

To assess stealthiness, we include a second question: ``Do you think this image is likely to appear in daily life?'' This question is designed to evaluate whether the image appears natural and inconspicuous, thereby avoiding potential bias that could arise from explicitly prompting participants to judge whether an image appears adversarial or malicious.

\textbf{Results.} Detailed results are presented in Table~\ref{tab:user-study}. As shown, our \system is generally considered a qualified appearance attack compared to other attacks with only 21\% of users perceiving our generated attack as a normal sign. Interestingly, NDD* (NDD attack containing the `STOP' text) had a high detection rate, with 68\% of human subjects identifying it as a STOP sign, suggesting that text is the primary factor in human recognition of traffic signs. As for stealthiness, our \system is more stealthy compared to FTE, NDD, and SIB, with 41\% of participants considering the generated images likely to appear in daily life: around three times higher than baselines. This increased stealthiness is due to our incorporation of elements, such as pandas, into the adversarial attacks, making them appear more realistic and similar to artwork in daily life. Although NDD* achieved high stealthiness with 63\% of users perceiving it as realistic, it is generally not qualified as an appearance attack, as 68\% of users recognized it as a STOP sign. Thus, among attacks that qualify as appearance attacks, \system demonstrates superior stealthiness.

\subsection{End-to-End Attack Impact Evaluation}
\label{sec:end2end}
\begin{figure}[!t]
\centering

{\includegraphics[width=1.0\linewidth]
{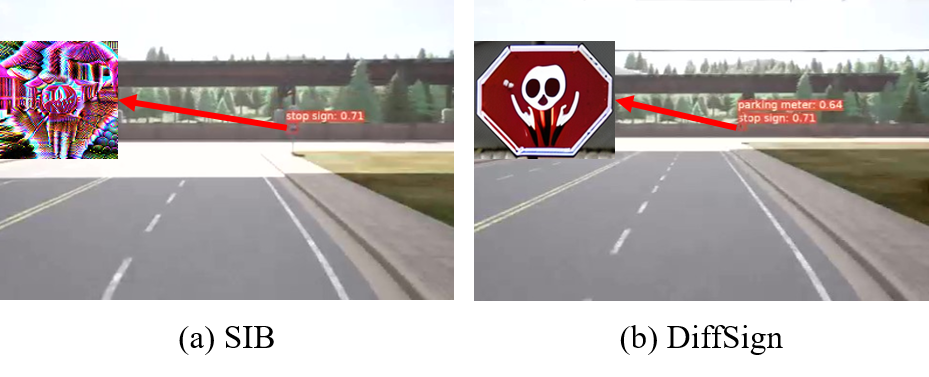}
}

\vspace{-0.4cm}
\caption{The frames illustrate the moment at which the adversarial appearance attack is first recognized by the vehicle’s TSR system during approaching. The left figure shows the SIB method~\cite{zhao2019seeing}, which only becomes effective at a short range, limiting its impact. In contrast, the right figure displays our \system, which is detected from a significantly greater distance. This extended effective range enables the vehicle to react earlier, leading to better end-to-end attack performance.}
%
\label{fig:e2e}

\end{figure}

{\bf Methodology and Setup.} To evaluate the end-to-end impact of our attack on AD systems such as its potential to induce unnecessary emergency stops, we utilize the \textit{PASS} platform~\cite{shen2022sok, hu2022pass}. \textit{PASS} has been widely adopted in AD security research for its representativeness and comprehensive simulation capabilities~\cite{Wang_2023_ICCV, ma2024slowtrack}. Our analysis focuses on the STOP sign scenario, a safety-critical context and a common target in TSR security studies~\cite{Wang_2023_ICCV, zhao2019seeing, eykholt2018physical, wang2025revisiting}. Simulations are conducted using the Carla simulator~\cite{Dosovitskiy17} with the default \textit{PASS} configuration~\cite{shen2022sok, hu2022pass}. We adopt the Attack Success Rate for STOP signs (ASR-S) as our evaluation metric, which measures the proportion of trials in which the vehicle performs a stop in response to an adversarial appearance attack. Each experiment is repeated 10 times with randomized initial vehicle positions to obtain a robust estimate of ASR-S, following the same setup in~\cite{Wang_2023_ICCV}. The scene is visualized in Fig.~\ref{fig:e2e}.

\begin{figure}[!t]
\centering

{\includegraphics[width=1.0\linewidth]
{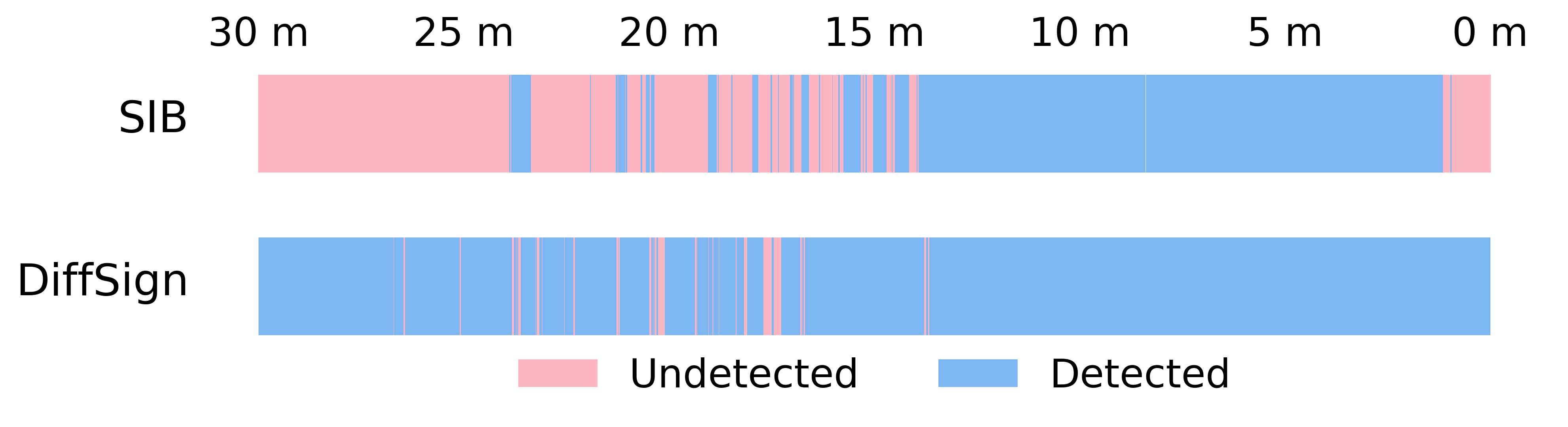}
}

\caption{Visualization of detection results of YOLO v3 comparing the SIB baseline and \system across a 0–30 meter range in the physical world. This setup represents a conservative scenario for \system but a favorable one for SIB, as YOLO v3 serves as the white-box attacked model for SIB, whereas \system operates under a more challenging black-box transfer attack setting, which generally yields lower effectiveness.}
\vspace{-0.2cm}
\label{fig:compare_distance}

\end{figure}

{\bf Results.} The results, summarized in Table~\ref{tab:system}. All appearance attacks are highly qualified appearance attacks based on our user study in~\S\ref{sec:attack_effectiveness}. Notably, our proposed \system achieves a 100\% ASR-S, whereas all baseline attacks yield an ASR-S of 0\%. The primary factor contributing to this performance gap is the effective attack range. As shown in Fig.~\ref{fig:e2e}, baseline methods such as SIB~\cite{zhao2019seeing} are only detected at short distances, providing insufficient time to influence the vehicle's decision-making process. In contrast, \system is recognized from a significantly greater distance, sustaining its effectiveness across a longer sequence of frames. To further illustrate the claim above, we compare the detection results of SIB and \system across a range of distances (0–30 meters) in the physical world, using the same experimental setup as described in our physical-world evaluations. For this comparison, we employ YOLOv3 as the object detector. This setup represents a conservative evaluation scenario for \system but a favorable one for SIB, as YOLO v3 serves as the white-box attacked model for SIB, whereas \system operates under a more challenging black-box transfer attack setting, which typically yields lower effectiveness. However, as shown in Fig.~\ref{fig:compare_distance}, \system demonstrates stable attack success rates from distances as far as 30 meters, while SIB only achieves effectiveness at approximately 15 meters. This extended effective range of \system is crucial for enabling end-to-end impact, such as triggering unnecessary emergency stops.

\begin{table}[t]
\small

    \caption{End-to-end attack impact evaluation with baselines. Each experiment is repeated 10 times with randomized initial vehicle positions. ASR-S denotes the proportion of trials in which the vehicle performs a stop in response to an adversarial appearance attack.}

    \centering
    \begin{tabular}{ccccc}

    \toprule
    & NDD & NDD* & SIB & \textbf{\system} \\
    \cmidrule(lr){2-5}
    & \includegraphics[height=1.3cm]{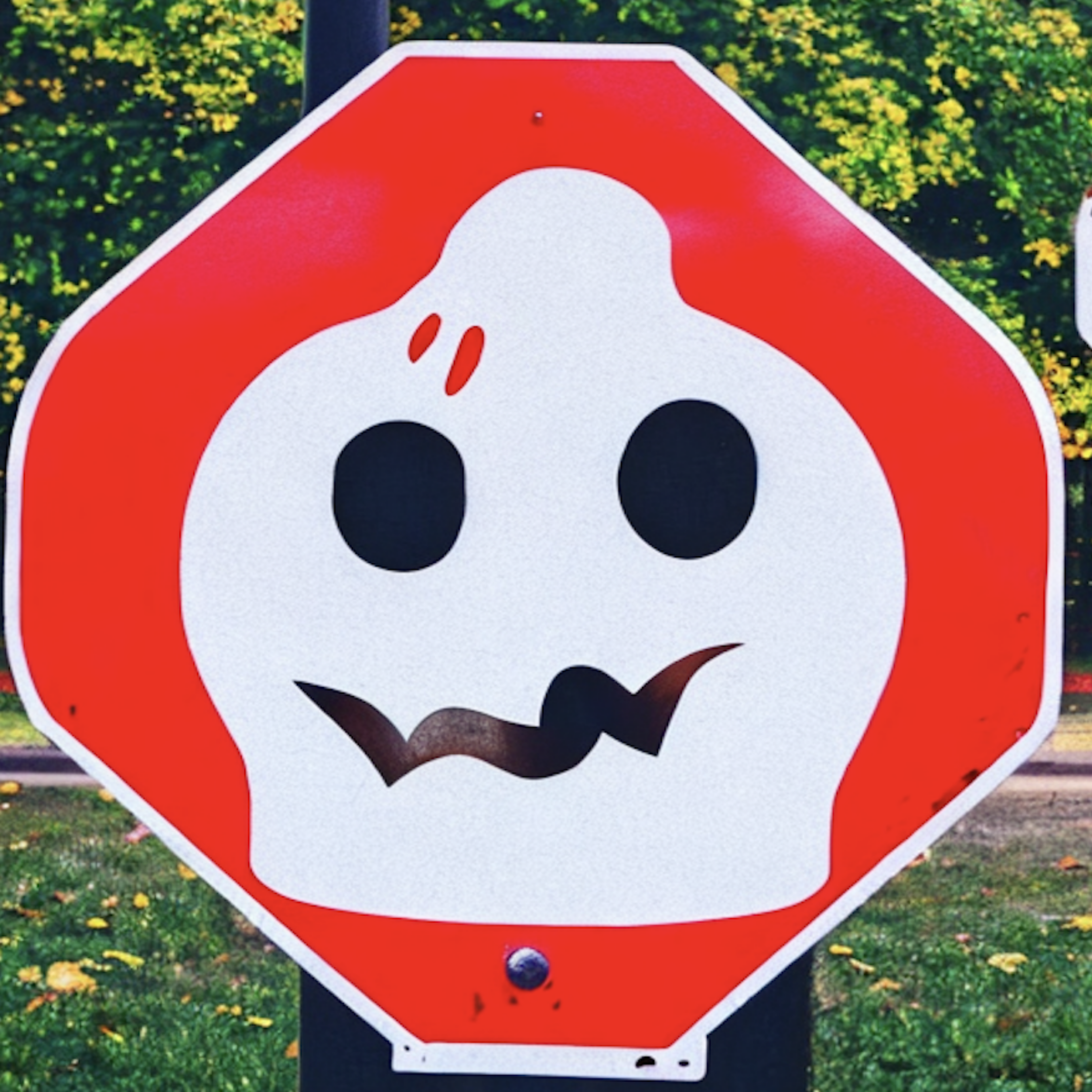}  & \includegraphics[height=1.3cm]{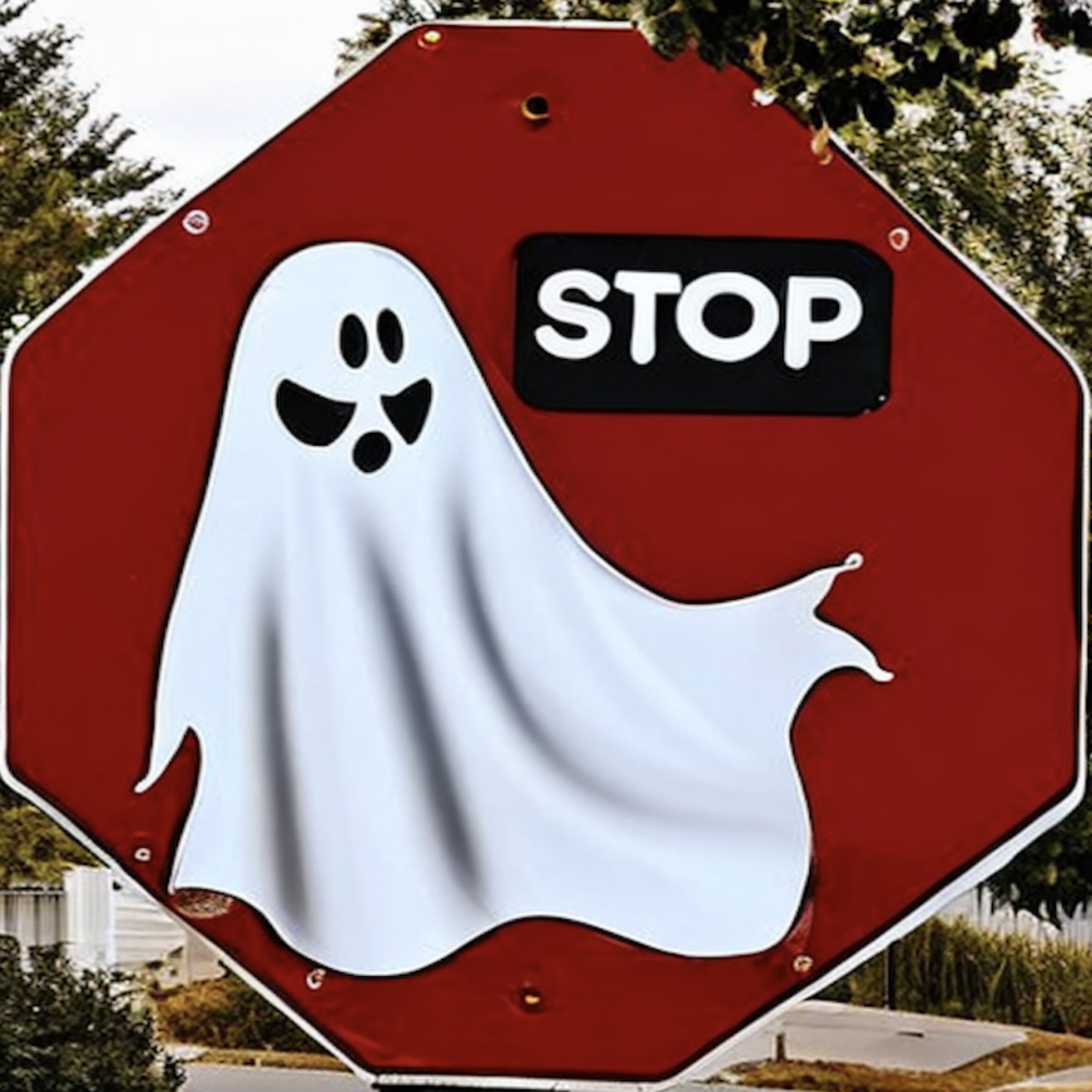}  & \includegraphics[height=1.3cm]{img/appear_to_stopsign.png}  & \includegraphics[height=1.3cm]{img/success_903_96.png}\\
    \midrule

    ASR-S & 0\% & 0\% & 0\% & \textbf{100\%} \\
    
    \bottomrule
         
    \end{tabular}

    \label{tab:system}
\end{table}

\begin{table}[t]
\tabcolsep 0.04in
\small

    \centering
    \caption{Ablation study evaluation for each design component of \system . w/o: without, w/: with, BF: BBOX Filter, I-C: Image-specified Customization, P-C: Prompt-specified Customization, C: Style Customization.}
    \label{tab:ablation}
    \begin{tabular}{ccccccccc}

    \toprule
     & w/o ${L}_{d}$  & w/o ${L}_{s}$ & w/o BF & w/ I-C & w/ P-C & w/o C\\
    \midrule
    STOP &0.9\% & 88.6\% & 19.2\% & 66.9\% & 85.8\% & 82.2\% \\
    \midrule
    Speed Limit & 0.3\% & 81.7\% & 22.4\% & 65.7\% & 82.5\% & 14.6\% \\

    \bottomrule
         
    \end{tabular}
    \vspace{-0.1cm}

\end{table}

\subsection{Ablation Study.}
\label{sec:ablation}
{\bf Methodology and Setup.} To evaluate the contribution of each design component to the overall effectiveness of our attack, we conduct an ablation study on both STOP signs and speed limit signs. Specifically, we systematically remove individual components—including ${L}_{s}$, ${L}_{d}$, and BBOX filtering; and assess the resulting impact on attack performance. For the style customization ablation study, we analyze three variants: (1) adding image-specific customization, (2) adding prompt-specific customization, and (3) removing style customization. The evaluation is conducted in digital simulation, following the same setup as detailed in~\S\ref{sec:Baseline}.

{\bf Results.} The results of the ablation study, presented in Table~\ref{tab:ablation}, reveal several key insights. The detection loss ${L}_{d}$ is critical to the success of the attack. When ${L}_{d}$ is removed, the adversarial effectiveness drops drastically, with the ASR falling to 0.9\% for STOP signs and 0.3\% for speed limit signs. In contrast, removing the similarity loss ${L}_{s}$ has minimal impact on ASR. However, we observed that the resulting adversarial examples often include visible text such as ``STOP'', thereby compromising the qualification of the attack. Additionally, the BBOX Filter plays an important role in eliminating cases shown in Fig.~\ref{fig:bad_case} (b), which can negatively affect attack performance.

Additionally, we assess the effectiveness of our style customization in the last three columns of Table~\ref{tab:ablation}. For object categories where text-to-image (T2I) diffusion models possess strong prior knowledge (e.g., STOP signs), style customization does not significantly improve ASR and may sometimes even slightly degrade performance. For instance, using image-specified customization results in 66.9\% ASR, whereas omitting it yields a higher ASR of 82.2\%. In such cases, the primary utility of style customization lies in its ability to generate adversarial examples with specific stylistic attributes aligned with attackers’ intent, rather than in significantly enhancing effectiveness.

In contrast, for out-of-domain categories such as Chinese speed limit signs, where T2I diffusion models lack inherent prior knowledge—style customization proves essential. Without it, relying solely on detection loss leads to poor guidance during generation, resulting in a low ASR of 14.6\%. However, with style customization, the ASR improves significantly. This indicates that our style customization techniques effectively inject prior knowledge into the generation process, addressing the T2I diffusion model's limitations in unfamiliar domains. Overall, these findings highlight that our methods not only enhance controllability but also improve the generalization capability of T2I-based adversarial attacks.

To sum up, our ablation study identifies two critical factors for enabling T2I diffusion models to achieve strong performance in appearance attacks: (1) effective guidance via detection loss, and (2) the incorporation of object-category-specific prior knowledge. The latter is effectively introduced through our style customization methods, which is important to unblock the potential of T2I diffusion models for appearance attacks.

\subsection{Different Prompts Comparison.}
{\bf Methodology and Setup.} To evaluate the effect of different prompt types and assess the relative importance of various traffic sign features for both TSR systems and human recognition, we construct variants by selectively removing key features—shape, color, text, and pattern—from the description of a standard STOP sign. Additionally, we include a prompt variant with all features removed. For each prompt type, we generate three different adversarial attacks. We manually inspect all generated images to ensure the specified feature is indeed removed while all other features remain unaffected. Following the same random digital simulation method introduced in~\S~\ref{sec:attack_effectiveness}, these attacks are placed into backgrounds for evaluation. We compute two metrics: ASR -- percentage of images in which the TSR model succeeds in recognizing the STOP sign, indicating a successful attack; and Human Detection Rate (HDR) -- We recruited 100 human participants and recorded the percentage of images where participants correctly recognized the image as a STOP sign.

{\bf Results.} Table~\ref{tab:prompt} presents the ASR and HDR results across different prompt types. For humans, the text on traffic signs is most critical for recognition, as evidenced by the significant drop in HDR when text is removed (down to 19\%). Conversely, TSR models rely more heavily on shape and color, as removing these features significantly increases ASR (62\% and 69\%, respectively). Text appears to be less important for TSR recognition, despite being essential for human perception. The findings highlight a divergence in perception between machines and humans.
These results demonstrate that \system enables semantic-level controllable generation, contrasting with prior pixel-level attacks~\cite{zhao2019seeing, jia2022fooling}. This controllability allows us to isolate and understand how different visual features influence recognition, exposing key discrepancies between model-learned and human-understood features—discrepancies that our attack successfully exploits.

\begin{table}[t]
\tabcolsep 0.1in
\small

    \caption{Different prompt types comparison on STOP sign case. HDR denotes Human Detection Rate.}
    \centering
    \begin{tabular}{cccccc}

    \toprule
    &  \multicolumn{5}{c}{Removed Feature }  \\
    \cmidrule(lr){2-6}
    & Shape & Color & Text & Pattern & All\\
    \midrule
    HDR & 35\% & 28\% &  19\% &  37\% &  16\% \\
    ASR & 62\% & 69\% &  82\% &  71\% &  54\% \\

    \bottomrule
         
    \end{tabular}

    \label{tab:prompt}
    \vspace{-0.3cm}
\end{table}

\subsection{Computation Costs and Time Analysis.} 
The hardware for our attack generation and evaluation is introduced in~\S\ref{sec:exp_setup}. \system takes 27 minutes to generate adversarial attacks, which is more efficient than traditional pixel-level perturbation attacks: 47 minutes for SIB~\cite{zhao2019seeing} and 63 minutes for FTE~\cite{jia2022fooling}. This advantage arises from traditional pixel-level methods that require numerous EoT iterations to improve adversarial attacks' physical-world robustness. In contrast, images generated by T2I diffusion models inherently possess strong physical-world robustness, as found in NDD~\cite{sato2024intriguing}. Therefore, our method's computational cost primarily stems from optimizing adversarial images within the T2I diffusion model's semantic space, without requiring significant additional time to improve physical-world robustness.
\section{Discussion}
\label{sec:discussion}

\begin{figure*}[!ht]
\centering
\includegraphics[width=0.9\linewidth]{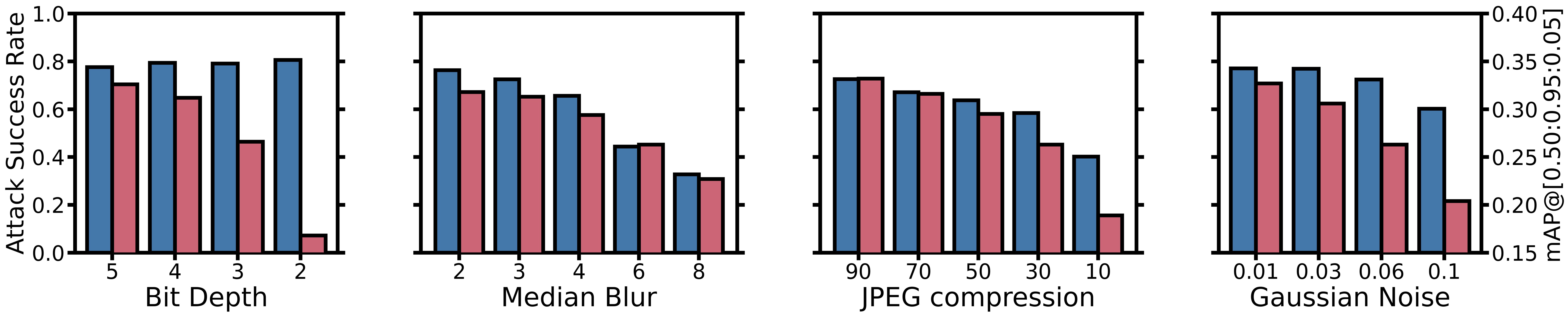}
\caption{Attack effectiveness regarding ASR (blue bars) and benign model performance regarding mAP (red bars) under four commonly used input-transformed defenses. The x-axis represents the strength of each defense.}
\label{fig:defense1}
\end{figure*}

\subsection{Defense}
\label{sec:defense}
We provide discussions of new challenges that our T2I-based attack \system poses to existing defenses.
Specifically, \system introduces two challenges: strong robustness under various image transformations, and high naturalness, which diminishes the effectiveness of defenses~\cite{wu2024napguard, liu2022segment, lin2024don} that rely on detecting unique features of adversarial patches. We consider general DNN-based defenses and domain-specific defenses. Furthermore, some promising defense directions are discussed.

{\bf General DNN-based Defenses.} We evaluate several common input-transformed defenses, which are directly adaptable with low computational overhead. These include JPEG compression~\cite{dziugaite2016study}, bit depth reduction~\cite{xu2017feature}, Gaussian noise~\cite{zhang2019defending}, median blur~\cite{xu2017feature}, and non-local means~\cite{xu2017feature, zhang2020interpretable}. These defenses can be applied to camera inputs before recognition, avoiding retraining and keeping inference delay within the real-time constraints of AD. Due to their easily adaptable nature, these methods have been assessed in recent security studies~\cite{cao2021invisible, sato2021dirty, zhu2023tpatch, zhang2020interpretable}.
The effectiveness of these defense measures is quantified by ASR, while the impact on benign performance is assessed using the mAP (mean Average Precision). The evaluation is performed on YOLO v5. As shown in Fig.~\ref{fig:defense1}, we observe that, under high defense strength, most defense mechanisms—excluding Bit Depth—are capable of partially mitigating the adversarial attacks. However, successful attacks can still occur, and when they do, they significantly degrade model performance. This vulnerability poses substantial risks in safety-critical applications, as even a single failure may lead to severe consequences\cite{zhu2023tpatch}. Therefore, despite their partial effectiveness, these defenses are not practically viable for real-world deployment.

\begin{figure}[!t]
\centering

{\includegraphics[width=0.8\linewidth]
{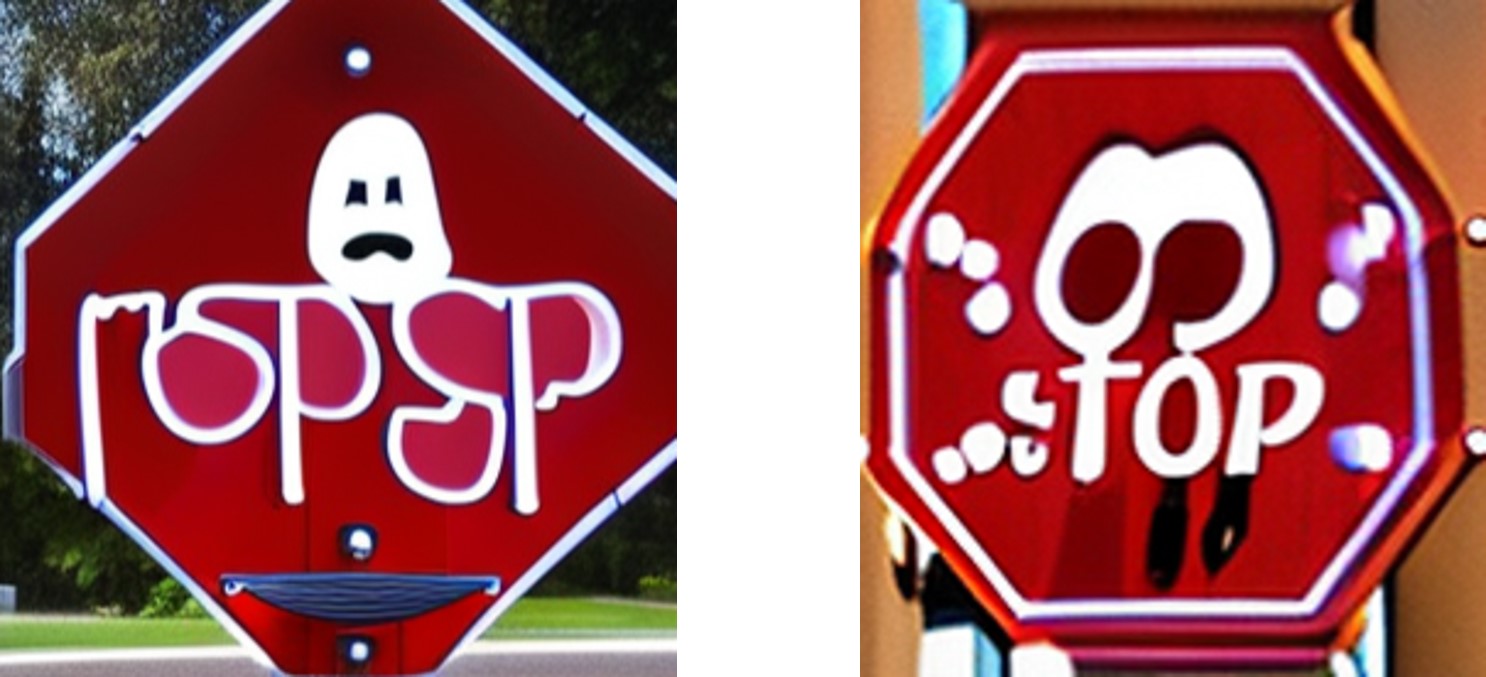}
}

\caption{Visualization of text-containing adversarial appearance attack that successfully bypasses ContraNet~\cite{Yang2022WhatYS}.}
\vspace{-0.3cm}
\label{fig:text_example}

\end{figure}

\textbf{Domain-specific Defenses.} 
Two representative types of domain-specific defenses are involved:  patch localization-based defenses, which detect and locate suspicious regions as adversarial patches; and consistency-based defenses, which identify attacks by measuring semantic inconsistency.

\textit{Patch Localization-based Defense.} We evaluate three patch localization-based defenses: NAPGuard~\cite{wu2024napguard}, which leverages an additional detection model to eliminate adversarial patches; SAC~\cite{liu2022segment}, which segments suspicious regions and masks them out; and NutNet~\cite{lin2024don}, which employs a generative model to detect adversarial content based on differences in data distribution between adversarial and clean images. Their ASR are 83\%, 58\%, and 44\%, respectively. Although NutNet appears to partially mitigate our attack, it is noteworthy that it reduces the TSR model’s detection rate for benign traffic signs to only 4\%, thereby severely degrading the model’s nominal performance. In contrast, while NAPGuard seems to exhibit the weakest defense effect, it minimally impacts the TSR model, maintaining a benign traffic sign detection rate of 97\%. This phenomenon can be attributed to the inherent trade-off between aggressive patch localization and preservation of clean-sample performance: defenses like NutNet, which strongly rely on generative reconstruction and distributional cues, may overfit to patch-like patterns and misclassify benign regions, whereas NAPGuard’s more conservative detection strategy prioritizes preserving clean-example accuracy even at the cost of reduced adversarial robustness. This limitation is due to the fact that our adversarial examples and benign examples exhibit minimal differences in both distribution and visual features, making it difficult for patch localization-based defenses to distinguish between them, resulting in either masking both or retaining both simultaneously. This stems directly from the high naturalness of our adversarial examples.

\textit{Consistency-based Defenses}. Another defense method, ContraNet~\cite{Yang2022WhatYS}, detects attacks using inconsistency between semantic information and discriminative features. We evaluate ContraNet on \system generated under different prompts. Our evaluation shows that while non-text adversarial attacks are successfully detected, text-containing examples are undetected. Some text-containing attacks that successfully bypass ContraNet are shown in Fig.\ref{fig:text_example}. This suggests that ContraNet pays more attention to textual features, which are more critical for human recognition of traffic signs compared to shape and texture features. Although text in these text-containing attacks may not be recognizable by humans, this detection method shows promising directions in attack mitigations.

{\bf Promising Defense Direction.} We believe that a promising defense strategy is to enhance the model's focus on textual features, forcing attackers to sacrifice attack qualification to bypass such defenses. Moreover, integrating stronger textual priors into the detection model can improve the robustness of TSR systems by reducing false positives from graffiti, advertisements, or unrelated textual content that might otherwise be misclassified as traffic signs. This not only strengthens defense against T2I-based appearance attacks but also improves system reliability in complex real-world environments.

\subsection{Limitation and Future Work}
Our study has several limitations that open avenues for future research. First, our physical-world evaluations are restricted to two commercial TSR systems, which may limit the generalizability of our findings to other vehicle models and TSR implementations~\cite{wang2025revisiting}. Future work could explore the effectiveness of our attack across a broader range of commercial systems and vehicle brands. Second, our evaluation focuses on two traffic sign types; extending this analysis to a wider variety of signs would further validate the generalizability and robustness of the proposed approach.

Moreover, while \system demonstrates strong performance across both open-source and commercial TSR systems, the underlying mechanisms driving its exceptional robustness and cross-model transferability remain poorly understood. Investigating why T2I-based adversarial attacks maintain such resilience in physical-world conditions—and why they transfer effectively across different architectures and environments—represents an important direction for future study.

Regarding baselines, SIB~\cite{zhao2019seeing} and FTE~\cite{jia2022fooling} serve as representative pixel-level appearance attack methods. For SIB, due to the unavailability of official source code, we utilized the adversarial attacks provided by the authors for evaluating attack effectiveness. Additionally, we re-implemented the example generation pipeline described in the original paper to conduct a time cost analysis. For FTE, we attempted to reproduce the attack following the methodology outlined in the paper; however, the absence of key implementation details—such as the transformation distributions and hyperparameter settings in the EoT process—precluded a fair comparison in terms of attack effectiveness. As such, we report only the time cost for FTE under our hardware configuration.

To promote transparency and enable future benchmarking, we will publicly release the complete source code for \system, along with all generated adversarial examples.

\section{Conclusion}
\label{sec:conclusion}

In this paper, we introduce \system, a novel T2I-based attack framework for generating appearance attacks which are physically robust, highly effective, transferable, and visually stealthy against TSR systems. \system incorporates new cropping mechanism, CLIP-based loss, and masked
prompts to improve attack focus and controllability.
Additionally, we introduce two style customization strategies that enable attackers to control the visual style of adversarial signs, thereby enhancing both stealthiness and generalization across different traffic sign types.
We conduct comprehensive real-world evaluations of \system across varying conditions, including changes in viewing distance, angle, lighting, and sign type.
Our method achieves an average physical-world attack success rate of 83.3\%, and reaches up to 97\% success on commercial TSR systems integrated into two top-selling production vehicles. We hope that our findings and insights can inspire future research into this critical domain.

\IEEEpeerreviewmaketitle
\bibliographystyle{IEEEtran}
%
\bibliography{main.bib}



\end{document}